\documentclass{article}
\usepackage{style/foci}
\usepackage{tgpagella}   
\usepackage{tgheros}     

\usepackage[utf8]{inputenc}
\usepackage[T1]{fontenc}
\usepackage{microtype}

\usepackage{amsmath}
\usepackage{amssymb}
\usepackage{amsfonts}
\usepackage{amsthm}
\usepackage{mathtools}
\usepackage{mathrsfs}
\usepackage{physics}
\usepackage{braket}
\usepackage{slashed}
\usepackage{nicefrac}
\usepackage{textcomp}
\usepackage{dsfont}
\usepackage{bbm}
\usepackage{bm}

\usepackage{graphicx}
\usepackage{subcaption}
\usepackage[export]{adjustbox}
\usepackage{float}
\usepackage{booktabs}
\usepackage{dcolumn}
\newcolumntype{d}[1]{D{.}{.}{#1}}
\usepackage{bigstrut, tabularx, multirow, makecell, diagbox}
\usepackage{colortbl}
\usepackage{tabularray}
\UseTblrLibrary{booktabs}
\usepackage{threeparttable}
\usepackage{tablefootnote}
\usepackage{fontawesome5}

\usepackage{placeins}
\usepackage{caption}
\usepackage{footnote}
\usepackage{enumitem}
\usepackage{multicol}
\usepackage{xspace}
\usepackage{titletoc}
\usepackage{titlesec}
\usepackage[bottom]{footmisc}
\usepackage{setspace}

\usepackage{wrapfig}
\usepackage{tikz}
\usepackage{quantikz}
\usepackage{dashbox}
\usepackage{mdframed}
\usepackage{marvosym}
\usepackage{pifont}
\usepackage{CJK}
\usepackage{url}

\usepackage[table,x11names]{xcolor}
\usepackage[most]{tcolorbox}
\tcbuselibrary{breakable}
\usetikzlibrary{decorations.pathreplacing, fit}

\definecolor{primaryblue}{HTML}{0066CC}
\definecolor{accentcyan}{HTML}{00D4AA}
\definecolor{warmorange}{HTML}{FF6B35}
\definecolor{deepgray}{HTML}{2C3E50}
\definecolor{lightgray}{HTML}{F8F9FA}
\definecolor{gradientstart}{HTML}{667eea}
\definecolor{gradientend}{HTML}{764ba2}

\definecolor{citecolor}{HTML}{0071bc}
\definecolor{citeblue}{RGB}{0, 113, 188}
\definecolor{linkcolor}{HTML}{9A4D92}
\definecolor{firebrick}{rgb}{0.698,0.133,0.133}

\definecolor{paleviolet}{HTML}{E1EEFC}
\definecolor{CarolinaUltraLight}{HTML}{E7F4FC}
\definecolor{lightgrey}{RGB}{247, 247, 247}
\definecolor{shadecolor}{HTML}{EFEFEF}
\definecolor{lightyellow}{rgb}{1.0, 0.95, 0.7}
\definecolor{lightblue}{rgb}{0.90, 0.95, 1.0}
\definecolor{light-gray}{gray}{0.95}

\definecolor{darkgrey}{rgb}{0.5, 0.5, 0.5}
\definecolor{darkgreen}{rgb}{0, 0.5, 0}
\definecolor{mydarkblue}{rgb}{0,0.08,0.45}
\definecolor{mydarkblue2}{rgb}{0.133, 0.133, 0.698}
\definecolor{echodrk}{HTML}{0099cc}
\definecolor{mymauve}{rgb}{0.58,0,0.82}
\definecolor{midnightblue}{rgb}{0.1,0.1,0.44}
\definecolor{oxfordblue}{rgb}{0.0,0.13,0.28}
\definecolor{prussianblue}{rgb}{0.0,0.19,0.33}
\definecolor{coolteal}{rgb}{0, 0.45, 0.45}
\definecolor{olive}{rgb}{0.1, 0.3, 0}
\definecolor{mypurple}{rgb}{0.5,0,0.5}
\definecolor{almond}{rgb}{0.94, 0.87, 0.8}

\definecolor{blue_ampEncoding}{HTML}{DAE8FC}
\definecolor{green_encoder}{HTML}{D5E8D4}
\definecolor{purple_decoder}{HTML}{E1D5E7}
\definecolor{yellow_measure}{HTML}{FFF2CC}
\definecolor{gray_block}{HTML}{F5F5F5}
\definecolor{pink_dru}{HTML}{FAD9D5}
\definecolor{orange_v}{HTML}{FAD7AC}

\definecolor{colorA}{rgb}{1,0,0}
\definecolor{colorB}{rgb}{0,0.3,1}
\definecolor{colorC}{rgb}{0.9,0.8,0.2}
\definecolor{colorD}{rgb}{0,0.65,0}
\definecolor{lesslightgray}{rgb}{0.5,0.5,0.5}
\definecolor{fundamental}{RGB}{55, 110, 111}
\definecolor{Gred}{RGB}{219, 50, 54}
\definecolor{ToCgreen}{RGB}{0, 128, 0}
\definecolor{Sepia}{RGB}{112, 66, 20}
\definecolor{Dblue}{rgb}{0,0.08,0.45}
\definecolor{Blue}{rgb}{0, 0, 0.8}
\definecolor{blue}{rgb}{0,0,1}
\definecolor{UNCblue!10}{rgb}{0.84,0.91,0.98}
\definecolor{RowAlt}{rgb}{0.98,0.98,0.99}

\definecolor{CarolinaBlue}{HTML}{7BAFD4}        
\definecolor{CarolinaLightBlue}{HTML}{B3D4E5}   
\definecolor{CarolinaUltraLight}{HTML}{E8F4F8}  
\definecolor{CarolinaText}{HTML}{1C2B33}        

\usepackage[pagebackref=true,breaklinks=true,colorlinks,hyperfootnotes=false]{hyperref}
\hypersetup{
  colorlinks,
  citecolor=citeblue,
  linkcolor=firebrick,
  urlcolor=firebrick
}
\usepackage[nameinlink,capitalize,noabbrev]{cleveref}

\titlespacing\section{0pt}{4pt plus 4pt minus 2pt}{-2pt plus 2pt minus 2pt}
\titlespacing\subsection{0pt}{2pt plus 4pt minus 2pt}{-2pt plus 2pt minus 2pt}
\titlespacing\subsubsection{0pt}{2pt plus 4pt minus 2pt}{-2pt plus 2pt minus 2pt}

\makeatletter
\def\th@remark{%
  \thm@headfont{\bfseries}%
  \normalfont 
  \thm@preskip\topsep \divide\thm@preskip\tw@
  \thm@postskip\thm@preskip
}
\makeatother

\theoremstyle{definition}

\tcbset{focithm/.style={
  enhanced, breakable, frame hidden,
  colback=black!5,
  width=\linewidth,
  enlarge left by=0pt, enlarge right by=0pt,
  boxsep=6pt, boxrule=0pt,
  left=4pt, right=4pt, top=1pt, bottom=1pt,
  arc=8pt,
  before skip=\topsep, after skip=\topsep
}}

\tcolorboxenvironment{theorem}{focithm}

\tcolorboxenvironment{lemma}{focithm}

\tcolorboxenvironment{corollary}{focithm}

\tcolorboxenvironment{proposition}{focithm}

\tcolorboxenvironment{definition}{focithm}

\tcolorboxenvironment{assumption}{focithm}

\tcolorboxenvironment{claim}{focithm}

\tcolorboxenvironment{problem}{focithm}

\tcolorboxenvironment{question}{focithm}

\newtcolorbox{titleblock}{
  enhanced,
  frame hidden,
  colback=CarolinaUltraLight,
  colframe=CarolinaUltraLight,
  boxrule=0pt,
  arc=10pt,
  left=14pt,
  right=14pt,
  top=14pt,
  bottom=14pt,
  width=\linewidth,
  before skip=12pt plus 4pt,
  after skip=12pt plus 4pt,
  grow to left by=1.5pt,
  grow to right by=1.5pt,
  before upper={
    \setlength{\parindent}{0cm}
    \setlength{\parskip}{0.5cm}
  }
}

\crefname{theorem}{Theorem}{Theorems}
\crefname{proposition}{Proposition}{Propositions}
\crefname{lemma}{Lemma}{Lemmas}
\crefname{corollary}{Corollary}{Corollaries}
\crefname{definition}{Definition}{Definitions}
\crefname{assumption}{Assumption}{Assumptions}
\crefname{remark}{Remark}{Remarks}
\crefname{problem}{Problem}{Problems}
\crefname{property}{Property}{property}
\crefname{question}{Question}{Questions}

\numberwithin{equation}{section}
\numberwithin{theorem}{section}
\numberwithin{proposition}{section}
\numberwithin{definition}{section}
\numberwithin{lemma}{section}
\numberwithin{assumption}{section}
\numberwithin{remark}{section}

\def\1{\bm{1}}

\makeatletter
\let\save@mathaccent\mathaccent
\newcommand*\if@single[3]{%
    \setbox0\hbox{${\mathaccent"0362{#1}}^H$}%
    \setbox2\hbox{${\mathaccent"0362{\kern0pt#1}}^H$}%
    \ifdim\ht0=\ht2 #3\else #2\fi
}
\newcommand*\rel@kern[1]{\kern#1\dimexpr\macc@kerna}
\newcommand*\widebar[1]{\@ifnextchar^{{\wide@bar{#1}{0}}}{\wide@bar{#1}{1}}}
\newcommand*\wide@bar[2]{\if@single{#1}{\wide@bar@{#1}{#2}{1}}{\wide@bar@{#1}{#2}{2}}}
\newcommand*\wide@bar@[3]{%
    \begingroup
    \def\mathaccent##1##2{%
        \let\mathaccent\save@mathaccent
        \if#32 \let\macc@nucleus\first@char \fi
        \setbox\z@\hbox{$\macc@style{\macc@nucleus}_{}$}%
        \setbox\tw@\hbox{$\macc@style{\macc@nucleus}{}_{}$}%
        \dimen@\wd\tw@
        \advance\dimen@-\wd\z@
        \divide\dimen@ 3
        \@tempdima\wd\tw@
        \advance\@tempdima-\scriptspace
        \divide\@tempdima 10
        \advance\dimen@-\@tempdima
        \ifdim\dimen@>\z@ \dimen@0pt\fi
        \rel@kern{0.6}\kern-\dimen@
        \if#31
        \overline{\rel@kern{-0.6}\kern\dimen@\macc@nucleus\rel@kern{0.4}\kern\dimen@}%
        \advance\dimen@0.4\dimexpr\macc@kerna
        \let\final@kern#2%
        \ifdim\dimen@<\z@ \let\final@kern1\fi
        \if\final@kern1 \kern-\dimen@\fi
        \else
        \overline{\rel@kern{-0.6}\kern\dimen@#1}%
        \fi
    }%
    \macc@depth\@ne
    \let\math@bgroup\@empty \let\math@egroup\macc@set@skewchar
    \mathsurround\z@ \frozen@everymath{\mathgroup\macc@group\relax}%
    \macc@set@skewchar\relax
    \let\mathaccentV\macc@nested@a
    \if#31
    \macc@nested@a\relax111{#1}%
    \else
    \def\gobble@till@marker##1\endmarker{}%
    \futurelet\first@char\gobble@till@marker#1\endmarker
    \ifcat\noexpand\first@char A\else
    \def\first@char{}%
    \fi
    \macc@nested@a\relax111{\first@char}%
    \fi
    \endgroup
    }
\makeatother
\let\bar\widebar

\DeclareMathAlphabet{\mathsfit}{\encodingdefault}{\sfdefault}{m}{sl}
\SetMathAlphabet{\mathsfit}{bold}{\encodingdefault}{\sfdefault}{bx}{n}

\DeclareMathOperator*{\argmin}{arg\,min}

\let\tilde\widetilde
\let\hat\widehat

\renewcommand{\arraystretch}{1.15}
\makeatletter

\def\foci@authors{}
\def\foci@affil{}
\def\foci@email{}
\def\foci@project{}
\def\foci@date{\today}
\def\foci@authorsep{}

\newcommand{\fociauthor}[2]{%
  \g@addto@macro\foci@authors{\foci@authorsep\mbox{#1\textsuperscript{#2}}\gdef\foci@authorsep{,\ }}}
\newcommand{\fociaffil}[1]{\gdef\foci@affil{#1}}
\newcommand{\fociemail}[1]{\gdef\foci@email{#1}}
\newcommand{\fociproject}[1]{\gdef\foci@project{#1}}
\newcommand{\focidate}[1]{\gdef\foci@date{#1}}
\newcommand{\fociauthorlist}{\gdef\foci@authorsep{}\foci@authors}
\newcommand{\focicorr}{\textrm{\Letter}}

\def\foci@blfootnote{\gdef\@thefnmark{}\@footnotetext}
\newcommand{\focifootnotes}{%
  \foci@blfootnote{\textsuperscript{*}Equal contribution.\quad
    \textsuperscript{\Letter}Corresponding author: \texttt{\foci@email}}}

\newlength{\foci@mw}
\newcommand{\focidrawmark}[3]{%
  \setlength{\foci@mw}{#2}%
  \begin{scope}[shift={#3}, x={\foci@mw/564.03}, y={-\foci@mw/564.03}]
    \foreach \cx/\cy in {282.02/66.03, 282.02/222.08, 215.77/348.8, 348.73/348.8,
                         414.88/474.93, 149.16/474.93, 67.06/222.08, 496.99/222.08}
      \fill[#1] (\cx,\cy) circle[radius=42.04];
  \end{scope}}
\newcommand{\focimark}[2][black]{%
  \setlength{\foci@mw}{#2}%
  \begin{tikzpicture}[x={\foci@mw/564.03}, y={-\foci@mw/564.03}]
    \useasboundingbox (0,0) rectangle (564.03,540.96);
    \focidrawmark{#1}{#2}{(0,0)}
  \end{tikzpicture}}

\fancypagestyle{focititlepage}{\fancyhf{}\renewcommand{\headrulewidth}{0pt}}

\newcommand{\foci@headerline}{%
  \noindent
  \begin{minipage}[c]{0.5\linewidth}
    \includegraphics[height=0.58cm]{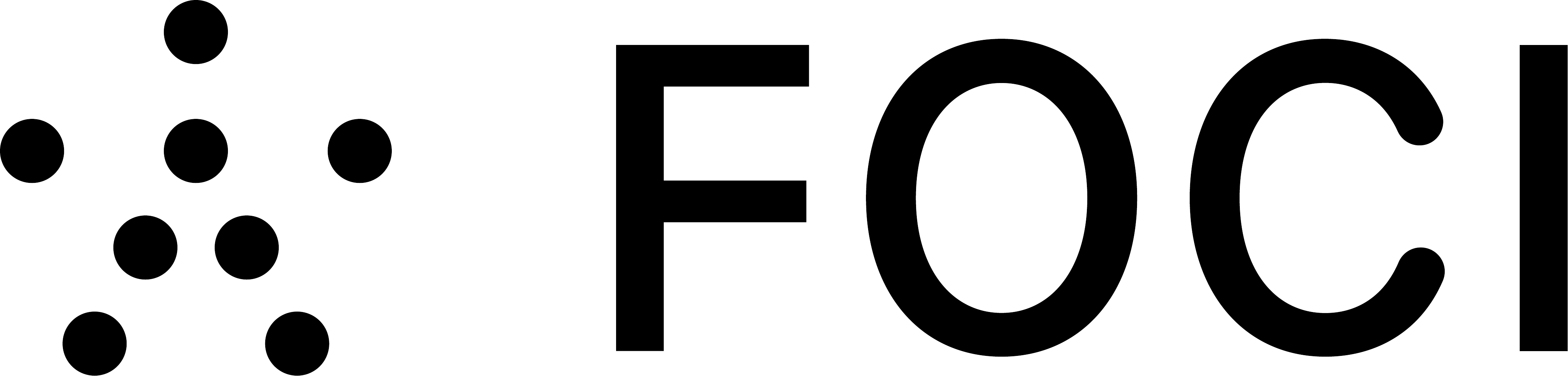}
  \end{minipage}%
  \begin{minipage}[c]{0.5\linewidth}
    \raggedleft\small\itshape\color{black!55}\foci@date
  \end{minipage}\par}

\newcommand{\foci@contactline}{%
  {\footnotesize\color{black!65}%
   \textbf{Email:}~\texttt{\foci@email}\qquad
   \textbf{Project:}~\url{\foci@project}\par}}

\newcommand{\foci@hairline}{\noindent\textcolor{black}{\rule{\linewidth}{0.5pt}}\par}

\newcommand{\focicoverCard}{%
  \thispagestyle{focititlepage}%
  \foci@headerline
  \vskip -0.5em
  \foci@hairline
  \vskip 0.9em
  \begin{tcolorbox}[enhanced, frame hidden, boxrule=0pt, arc=12pt,
      colback=black!4, width=\linewidth,
      left=18pt, right=18pt, top=18pt, bottom=16pt,
      grow to left by=2pt, grow to right by=2pt]
    \noindent
    \begin{minipage}[t]{\linewidth}
      {\setstretch{1.15}\LARGE\bfseries\raggedright LineupRL: Verifiable Reinforcement Learning for Time Series Captioning via Caption-to-Series Identification
\unskip\par}
      \vskip 0.55em
      {\normalsize\fociauthorlist\par}
      \vskip 0.2em
      {\small\color{black!70}\foci@affil\par}
    \end{minipage}\par
    \vskip 0.9em
    {\small\setstretch{1.08}Time series captioning is a fundamental step in time series understanding and
can also serve as the bridge between signal and natural language.
Supervised fine-tuning (SFT) relies on a larger
model's captions and cannot exceed their quality.
Reinforcement learning (RL) can, but its rewards were designed for other
modalities and other tasks, and they transfer poorly to open-ended generation in the time
series domain.
We address this by proposing \method, a reinforcement learning with verifiable rewards (RLVR)
pipeline whose reward is caption-to-series identification.
The reward model is a frozen large language model (LLM) verifier that reads the generated
caption and the candidate time series as raw values, never the chart, and must pick the
described time series from multiple distractors.
Matching is a far lighter demand on the verifier than writing questions or judging a caption, so an off-the-shelf LLM can supply the reward.
Across two captioning benchmarks, and on
forecasting and reconstruction where the predictor sees only the caption, \method\
outperforms SFT and RL baselines on every metric.
The 3B vision language model (VLM) trained by \method\ also outperforms, at $1/24$ of the
parameters, the 72B VLM whose captions the SFT baseline is distilled from.
Our case study shows that \method\ resists reward hacking, and that the captioner it trains both
traces the trend and names the values at key points.
\unskip\par}
    \vskip 0.9em
    \foci@contactline
  \end{tcolorbox}
  \vskip 0.4em}

\newcommand{\focicover}{\focicoverCard}

\makeatother

\usepackage{array}
\usepackage[normalem]{ulem}   
\usepackage{needspace}        

\newenvironment{promptbox}{%
  \begin{quote}\footnotesize\ttfamily\frenchspacing\raggedright
  \setlength{\parindent}{0pt}\setlength{\parskip}{0pt}\obeylines}{\end{quote}}

\newcommand{\method}{LineupRL\xspace}

\renewcommand{\ev}[1]{\uline{#1}}

\makeatletter
\renewcommand{\focifootnotes}{%
  \foci@blfootnote{\textsuperscript{\Letter}Corresponding author: \texttt{\foci@email}}}
\makeatother

\begin{document}

\raggedbottom

\fociauthor{Haochen Zhang}{1}
\fociauthor{Laura Yao}{1}
\fociauthor{Zachary Plotkin}{2}
\fociauthor{Gengwei Zhang}{1}
\fociauthor{Tianlong Chen}{1,2\,\focicorr}

\fociaffil{\textsuperscript{1}University of North Carolina at Chapel Hill\quad
  \textsuperscript{2}Foci Labs}
\fociemail{tianlong@cs.unc.edu}
\fociproject{https://github.com/HaochenZhang717/LineupRL-official-repo}

\pagestyle{fancy}
\fancyhf{}
\renewcommand{\headrulewidth}{0.4pt}
\chead{\small\scshape LineupRL: Verifiable Reinforcement Learning for Time Series Captioning via Caption-to-Series Identification
\unskip}
\cfoot{\small\thepage}

\focicover
\focifootnotes

\section{Introduction}
\label{sec:intro}
\begin{wrapfigure}[22]{r}{0.38\textwidth}
    \vspace{-12pt}
    \centering
    \includegraphics[width=0.9\linewidth]{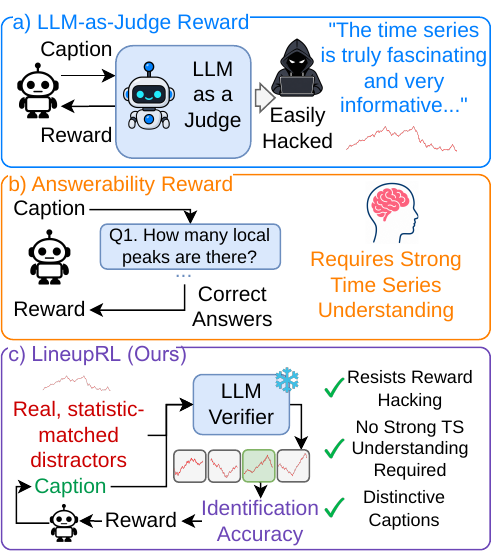}
    \caption{(a) LLM-as-judge is easily hacked. (b) Answerability needs strong time series
    understanding. (c) \method\ (ours).}
    \label{fig:intro}
\end{wrapfigure}

Time series record how a system evolves over time, and decisions in healthcare, finance and climate
depend on reading them accurately. Time series captioning automates this reading. Given a time
series, a captioner writes a natural-language description of how the signal behaves
\citep{jhamtani2021truth,sen2025bedtime,zhou2026catsbench}. Time series captions are useful from
multiple perspectives. For human practitioners, text is easier to act on than the raw
signal. For example, neonatal intensive-care staff chose better treatments from textual summaries of
physiological time series than from the corresponding trend graphs \citep{law2005comparison}. For
language-oriented systems, a caption provides a text interface to the signal, and this interface has been
used for domain reporting \citep{fons2024tadacap}, forecast explanation \citep{aksu2024xforecast}
and clinical prediction \citep{ji2026portable}. The
task dates back to rule-based data-to-text systems
\citep{kukich1983design,reiter2005choosing,yu2007choosing,portet2009automatic} and later data-driven
models \citep{lloyd2014automatic,jandaghi2020human,jhamtani2021truth}, both confined to fixed rules
or output formats. Large language models (LLMs), vision-language models (VLMs) and
time-series-language models (TSLMs) remove this restriction by open-ended generation, and recent
work improves their captions by supervised fine-tuning (SFT) on time series paired with captions. Because
human-written captions exist only in small collections \citep{jhamtani2021truth}, these pairs are
synthesized, either by prompting a larger model to describe the time series
\citep{trabelsi2025tslm,lin2024tessa} or by generating the text from attributes known by
construction \citep{xie2024chatts,feng2026decoupling}.

However, SFT has two weaknesses in this setting. First, it is bounded
by the supervised data. A captioner trained to imitate synthesized captions can be no better than the model
that wrote them, and it inherits that model's disadvantages, including plausible but incorrect statements
about the time series \citep{trabelsi2025tslm}. Second, it generalizes poorly. Trained on a limited
number of pairs, the captioner may memorize the reference captions rather than learning to read the
time series, and what it memorized transfers poorly to time series it has not seen \citep{chu2025sft}. On CaTS-Bench, open-source models fine-tuned on its synthetic
pairs still trail proprietary models on numeric detail, and the fine-tuned model states statistics
it cannot compute as confident guesses, which the benchmark attributes to a cross-entropy objective
that treats number tokens like any other token \citep{zhou2026catsbench}. In our own experiment,
fine-tuning a 3B VLM on captions from a 72B teacher does not raise its entailment of the reference
captions on BEDTime and lowers its series-to-caption identification on CaTS-Bench below the
untuned initialization (Table~\ref{tab:criteria12}). We call this failure mode
ungrounded captioning. The caption is fluent and plausible, but it misses the important
features of the time series, because the captioner learned the style of its references instead of
learning to read the series.

Reinforcement learning (RL) addresses both weaknesses. Trained with a verifiable reward, the
captioner learns from its own samples rather than from demonstrations, so no paired data is needed
\citep{lambert2024tulu,guo2025deepseekr1}. RL also generalizes where SFT memorizes
\citep{chu2025sft}, in part because its updates stay close to the base model
\citep{shenfeld2025razor}. However, RL needs a good reward design, and the representative rewards used for image
captioning do not work on time series (Figure~\ref{fig:intro}).
An LLM-as-judge reward (Figure~\ref{fig:intro}a) takes an LLM's grade of the caption as the
reward. Because the grade is not verifiable, it is easily hacked
\citep{xing2025caprl}. We observe that the captioner learns to game the grader instead of learning to describe
the time series. An answerability reward
(Figure~\ref{fig:intro}b) asks whether questions about the time series can be answered from the caption
alone \citep{xing2025caprl}. However, writing good questions and correct answers requires strong time series
understanding, which current models largely lack \citep{zhou2026catsbench,sen2025bedtime}.
A reward for time series captioning therefore has to satisfy three
conditions. It must (i) resist reward hacking; (ii) demand only as much time series understanding as current LLMs or
VLMs can supply; and (iii) reward important and distinctive features in time series. 

In this work, we propose \method, a reinforcement learning with verifiable rewards
(RLVR) pipeline built on caption-to-series identification, a reward that satisfies
all three desiderata (Figure~\ref{fig:intro}c).
Our starting point is that a caption is grounded and specific if a reader who sees only the
caption can pick the described time series out of close alternatives.
\method\ renders each time series as a line chart and trains a VLM
captioner on it, so the input does not grow with the time series length. For each training time series it
draws $K-1$ distractors, which are real time series of the same length that are nearest to it in summary
statistics. A frozen LLM verifier receives the caption and the $K$ candidates as raw values and names
the time series corresponding to the caption. The reward is the fraction of placements on which it names the
true time series (Figure~\ref{fig:intro}c), and each of these choices answers one of the three conditions.
(i) The reward is an
accuracy reward, the exact match of the verifier's pick against the ground-truth index, rather than
a judged score of the kind that is prone to reward hacking \citep{guo2025deepseekr1}.
(ii) The verifier solves a $K$-way multiple-choice question over printed numbers, which
needs matching rather than question writing or grading, and a 14B off-the-shelf LLM supplies it.
(iii) The $K-1$ distractors are selected to concur with the true time series on general statistics, so the caption must describe the finer shape to earn the reward.
Our contributions are as follows:

\begin{itemize}[leftmargin=*,nosep]
\item We trace ungrounded captioning to behavioral cloning of synthesized references, show why the
two existing families of RL rewards for captioning fail on time series, and state the three conditions a
verifiable reward for open-ended captioning must satisfy.
\item We propose \method, whose reward is caption-to-series identification by a frozen verifier
among statistic-matched distractors, which needs no reference captions.
\item We evaluate captions by entailment, identification and sufficiency for forecasting and
reconstruction, and show that \method\ improves all three at once and that a 3B captioner trained this way surpasses
the 72B teacher of the SFT baseline.
\end{itemize}

%
\section{Method}
\label{sec:method}

\subsection{Preliminaries}
\label{sec:method:setup}
We work with domain-agnostic univariate time series
$x=(x_1,\dots,x_T)\in\mathbb{R}^{T}$, where the length $T$ varies
across items.
Time series are drawn from a data distribution $\mathcal{D}$, written $x\sim\mathcal{D}$.
A renderer $\rho$ converts a time series into a line chart $\rho(x)$ that carries the curve including some axis ticks without per-point value labels.
A VLM captioner $\pi_\theta$ reads the chart $\rho(x)$ and outputs a natural-language
description $c\;\sim\;\pi_\theta\big(\cdot\mid\rho(x)\big)$, $\theta$ is updated during training so that $\pi_\theta$ produces captions with better quality.
Caption quality is measured by three types of criteria (\S\ref{sec:method:eval}).
The first type of criteria measures whether the generated caption implies a reference caption.
The second measures caption-to-series identification and
series-to-caption identification.
The third treats the caption as an information bottleneck on the
values and measures reconstruction and forecasting error.

\textbf{Notations.}
Throughout, $K\ge 2$ is the number of candidates in every identification task of this paper,
and $[K]=\{1,\dots,K\}$. Let $\varphi(x)=(\mathrm{mean},\mathrm{std},\min,\max)$ be the summary
statistics of a series. For each $x$, $\mathcal{N}(x)=\{\tilde{x}_1,\dots,\tilde{x}_{K-1}\}$
denotes the distractor set used in training, constructed in \S\ref{sec:method:ours}, and
$\mathcal{C}(x)=\{x\}\cup\mathcal{N}(x)$ the training candidate set. Their evaluation
counterparts, $\bar{x}_1,\dots,\bar{x}_{K-1}$ and
$\tilde{\mathcal{C}}(x)=\{x,\bar{x}_1,\dots,\bar{x}_{K-1}\}$, are constructed in
\S\ref{sec:method:eval}. A verifier is an LLM that receives a caption together with the $K$
candidates and answers this $K$-way multiple-choice question with a position in $[K]$. Because
the candidates are presented to a verifier as a list, we write $O_p(\mathcal{C}(x))$ for the
list in which the distractors appear in a fixed order and $x$ is inserted at position
$p\in[K]$, and $O_p(\tilde{\mathcal{C}}(x))$ for the same construction on the evaluation
candidates. $R$ denotes the verifier used in training, so $R\big(c,O_p(\mathcal{C}(x))\big)\in[K]$
is the position of the series that $R$ believes the caption $c$ describes. Evaluation uses a
single frozen LLM, which need not be the LLM behind $R$, in two roles:
$\tilde{R}_{\mathrm{c}\to\mathrm{s}}$ locates a series from a caption, and
$\tilde{R}_{\mathrm{s}\to\mathrm{c}}$ locates a caption from a series.

\textbf{RL with verifiable rewards.}
We post-train the captioner with reinforcement learning from verifiable rewards (RLVR)
\citep{lambert2024tulu,guo2025deepseekr1}. In this paradigm, the policy $\pi_\theta$ generates
captions, each generated caption is scored by a reward function, and reinforcement learning
updates $\theta$ to raise the probability of generations with higher rewards. Formally, given a
reward function $r(c, x)$ that scores a caption $c$
written for a series $x$, training maximizes the expected reward
$J(\theta)=\mathbb{E}_{x\sim\mathcal{D}}\,\mathbb{E}_{c\sim\pi_\theta(\cdot\mid\rho(x))}[r(c,x)]$.
We optimize $J(\theta)$ with the REINFORCE leave-one-out estimator (RLOO)
\citep{kool2019buy,ahmadian2024back}. Each training step draws a batch
$\{x^{(1)},\dots,x^{(B)}\}$ of $B$ series independently from $\mathcal{D}$. For each $b\in[B]$,
the step samples a group of $G$ captions $c^{(b)}_1,\dots,c^{(b)}_G$ independently from
$\pi_\theta\big(\cdot\mid\rho(x^{(b)})\big)$ and scores each caption with the reward. Since
the attainable reward varies with the series, caption $c^{(b)}_i$ is credited by the
leave-one-out contrast
$A^{(b)}_i=r\big(c^{(b)}_i,x^{(b)}\big)-\frac{1}{G-1}\sum_{j\neq i}r\big(c^{(b)}_j,x^{(b)}\big)$,
$i\in[G]$, whose subtracted term estimates the mean reward of $x^{(b)}$ from its other
$G-1$ captions and is therefore a variance-reducing baseline. The step then updates
$\theta$ by ascending the policy-gradient estimate
$\hat{g}=\frac{1}{BG}\sum_{b=1}^{B}\sum_{i=1}^{G}
A^{(b)}_i\,\nabla_\theta\log\pi_\theta\big(c^{(b)}_i\mid\rho(x^{(b)})\big)$ of
$\nabla_\theta J(\theta)$. The reward $r$ is the design choice of \S\ref{sec:method:ours}.

\subsection{\method}
\label{sec:method:ours}

\begin{figure}[t]
    \centering
    \includegraphics[width=\textwidth]{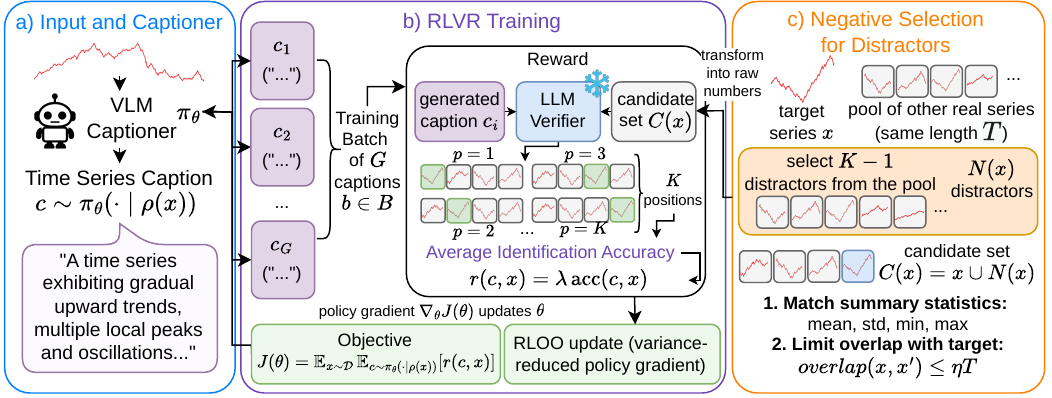}
    \caption{Overview of \method. (a) The VLM captioner $\pi_\theta$ reads the chart
    $\rho(x)$ of a series $x$ and samples a caption $c$. (b) RLVR training. For each series a
    group of $G$ captions is sampled. Each caption is scored by a frozen LLM verifier that receives
    the caption and the candidate set $\mathcal{C}(x)$ printed as raw numbers and must locate
    $x$; the true series is placed at each of the $K$ positions in turn, and the fraction of
    correct answers gives the reward $r(c,x)=\lambda\,\mathrm{acc}(c,x)$. The rewards of the
    group drive the RLOO policy-gradient update of $\theta$. (c) Negative selection. The $K-1$
    distractors $\mathcal{N}(x)$ are drawn from a pool of other real series of the same length by
    matching summary statistics and limiting overlap with $x$, and
    $\mathcal{C}(x)=\{x\}\cup\mathcal{N}(x)$.}
    \label{fig:method}
\end{figure}

Figure~\ref{fig:method} gives an overview of \method, which instantiates the RLVR of
\S\ref{sec:method:setup} with the captioner $\pi_\theta$ as the policy and RLOO as the
policy-gradient estimator; the component specific to \method\ is the reward. The reward is
caption-to-series identification (upper box of Figure~\ref{fig:method}b): it scores a caption
$c$ written for a series $x$ by whether the frozen verifier $R$ can identify $x$ in
$\mathcal{C}(x)$ from $c$ alone. The verifier sees only the caption and the $K$ candidates
printed as raw numbers, without seeing the chart $\rho(x)$, because the weak chart comprehension
of a VLM verifier would add noise to the reward. Since a single query at one fixed position
would be noisy and positionally biased, we query the verifier $K$ times, once with $x$ at each
position $p\in[K]$, and take the fraction of correct answers,
$\mathrm{acc}(c,x)=\frac{1}{K}\sum_{p=1}^{K}
\mathbf{1}\big[R\big(c,O_p(\mathcal{C}(x))\big)=p\big]$.
The reward is the rescaled accuracy $r(c,x)=\lambda\,\mathrm{acc}(c,x)$ with $\lambda=2$. A
caption therefore earns reward only through statements that separate $x$ from its neighbors.
Text addressed to the verifier or a generic description changes no answer, and a fabricated
value that happens to match a distractor sends the verifier to the wrong candidate, so only
grounded and distinctive content is rewarded.

The quality of the reward is largely determined by the distractor set $\mathcal{N}(x)$, whose
construction is shown in Figure~\ref{fig:method}c. 
A straightforward construction, deriving the
distractors from $x$ itself by shuffling, reversing, or cutting and pasting, is inappropriate
because such distractors are easily detected by the LLM verifier: under this construction the
verifier identifies $x$ at $72.5\%$ even without reading the caption. To avoid such information
leakage, \method\ draws $\mathcal{N}(x)$ from a pool of other real series of the same length as
$x$, and makes the question hard by two criteria, matching summary statistics and limiting
overlap with $x$. Let $\mathit{overlap}(x,x')$ be the longest overlap the two series $x$ and $x'$
share. The distractor set is
$\mathcal{N}(x)=\argmin_{\mathcal{S}}\sum_{x'\in\mathcal{S}}\|\varphi(x)-\varphi(x')\|_2$,
where $\mathcal{S}$ ranges over sets of $K-1$ other corpus series of length $T$ with
$\mathit{overlap}(x,x')\le\eta T$ for all $x'\in\mathcal{S}$, and $\eta\in(0,1)$ is an overlap budget. In other words, $\mathcal{N}(x)$
collects the $(K-1)$ nearest neighbors of $x$ under $\varphi$ among series of the same length,
excluding any that overlaps $x$ on more than a fraction $\eta$ of its steps. In training we
apply the same rule to the eight nearest admissible neighbors of $x$, which form a pool from
which $K-1$ distractors are redrawn at every step so that the caption cannot overfit one fixed
$\mathcal{N}(x)$; validation items keep one fixed distractor set. The candidate set
$\mathcal{C}(x)=\{x\}\cup\mathcal{N}(x)$ is then handed to the verifier. When candidates match
the target in length and summary statistics, the verifier must rely on the fine-grained
description to distinguish them.

\subsection{Evaluation protocol}
\label{sec:method:eval}

Captions are generated by greedy decoding at inference, so each series $x$ has one caption
$c(x)$, and every evaluation criterion is an empirical average over an evaluation set
$\mathcal{E}=\{x^{(1)},\dots,x^{(N)}\}$.
We apply three criteria, whose implementation details are given in Appendix~\ref{app:eval}.

\textbf{Criterion 1: entailment against reference captions.} Let $H(x)$ be the set of
reference captions of $x$, and let $q(a,b)\in[0,1]$ be the probability that a fixed
natural-language inference model assigns to the entailment class for premise $a$ and hypothesis
$b$. Over the pairs $\mathcal{P}=\{(x,h):x\in\mathcal{E},\,h\in H(x)\}$ we report
$\widehat{\mathcal{A}}=|\mathcal{P}|^{-1}\sum_{(x,h)\in\mathcal{P}}\mathbf{1}[\,q(c(x),h)\ge\tau\,]$,
the fraction of reference captions that the generated caption implies, where $\tau\in(0,1)$
is a fixed decision threshold. The generated caption is
the premise and the reference caption is the hypothesis, so $\widehat{\mathcal{A}}$ asks
whether the content of a reference caption is recoverable from the generated one while allowing
the generated caption to say more; the reverse direction is not reported.

\textbf{Criterion 2: identification by a frozen verifier.} This criterion runs identification
in two directions:
in caption-to-series identification (\textsc{cap}$\rightarrow$\textsc{ser}) the verifier receives a caption and $K$ candidate series
and must locate the described series, and in series-to-caption identification (\textsc{ser}$\rightarrow$\textsc{cap}) the verifier
receives a series and $K$ candidate captions and must locate the caption written for it.
Both directions use the evaluation candidate set $\tilde{\mathcal{C}}(x)$, whose distractors $\bar{x}_1,\dots,\bar{x}_{K-1}$ are the $K-1$
nearest neighbors of $x$ under $\varphi$ among evaluation series of the same length;
$\tilde{\mathcal{C}}(x)$ is computed from the series alone and reads no caption, so it is
identical for different captioners.
\textsc{cap}$\rightarrow$\textsc{ser} is the same caption-to-series identification task that the reward of
\S\ref{sec:method:ours} is built on, except that the verifier is
$\tilde{R}_{\mathrm{c}\to\mathrm{s}}$ and the candidates come from $\tilde{\mathcal{C}}(x)$.
Its identification accuracy is
$\widetilde{\mathrm{acc}}_{\mathrm{c}\to\mathrm{s}}(c,x)=\frac{1}{K}\sum_{p=1}^{K}
\mathbf{1}\big[\tilde{R}_{\mathrm{c}\to\mathrm{s}}\big(c,O_p(\tilde{\mathcal{C}}(x))\big)=p\big]$,
and we report its average over the evaluation data,
$\widehat{\mathcal{M}}_{\mathrm{c}\to\mathrm{s}}
=|\mathcal{E}|^{-1}\sum_{x\in\mathcal{E}}\widetilde{\mathrm{acc}}_{\mathrm{c}\to\mathrm{s}}(c(x),x)$.
In \textsc{ser}$\rightarrow$\textsc{cap}, the verifier receives the series $x$ and $K$
candidate captions, and must locate the caption written for $x$. The candidate captions are
those the same captioner wrote for the members of $\tilde{\mathcal{C}}(x)$. For $p\in[K]$, let
$Q_p(x):=(c(\bar{x}_1),\dots,c(\bar{x}_{p-1}),c(x),c(\bar{x}_p),\dots,c(\bar{x}_{K-1}))$
denote the ordered caption list with $c(x)$ inserted at position $p$; it is the caption-side
counterpart of $O_p(\tilde{\mathcal{C}}(x))$. The identification accuracy again averages over
all $K$ placements of $c(x)$ to avoid the verifier's position bias,
$\widetilde{\mathrm{acc}}_{\mathrm{s}\to\mathrm{c}}(x)=\frac{1}{K}\sum_{p=1}^{K}
\mathbf{1}\big[\tilde{R}_{\mathrm{s}\to\mathrm{c}}\big(x,Q_p(x)\big)=p\big]$,
and we report its average over the evaluation data,
$\widehat{\mathcal{M}}_{\mathrm{s}\to\mathrm{c}}
=|\mathcal{E}|^{-1}\sum_{x\in\mathcal{E}}\widetilde{\mathrm{acc}}_{\mathrm{s}\to\mathrm{c}}(x)$.

\textbf{Criterion 3: sufficiency as an information bottleneck.} The last criterion asks how much
of the series itself survives the encoding into language. We fit a decoder that reads the caption and predicts a numeric target on the chronologically earliest windows of each corpus, and report its error $\widehat{\mathcal{S}}$ on the latest, so that no test target precedes any training window. Two targets are
used: the horizon that follows the described window (forecasting) and the window itself
(reconstruction). Because the series is never an input to the decoder, only what the caption states
can lower $\widehat{\mathcal{S}}$. A low error is necessary but not sufficient, since a caption
that only names a pattern common to the dataset would score well while identifying nothing, so we
report it against three references: a text-blind predictor $b$ that never reads a caption, the
decoder refitted on an empty caption, and a detachment control in which the decoder is refitted
with every window paired with another window's caption. A caption counts as informative only if
its error stays far below these.
\section{Experiments}
\label{sec:exp}

\subsection{Experimental setup}
\label{sec:exp:setup}

We train on $10{,}000$ fragments sampled from TSFragment-600K \citep{ge2025t2s} under a fixed
seed, spread roughly evenly over the three lengths. Charts keep a handful of $x$ and $y$ tick
labels, so the sampling positions and the value range remain readable, but carry no title,
legend, series name, unit or per-point value labels, so the policy cannot rely on the dataset
name or read values off the chart, and has to describe the shape. Every
identification task uses $K=4$ candidates. Criteria 1 and 2 are measured on BEDTime and
CaTS-Bench; since the CaTS-Bench references name dates, places, domains and exact values that
a bare chart does not carry, Criterion 1 on CaTS-Bench scores against references that Claude
Opus 5 rewrote to remove them (Appendix~\ref{app:cats-rewrite}). Criterion 3 is measured on
four forecasting corpora (ETTh2, ETTm2, Saugeen river flow and Australian electricity demand)
under the chronological split of \S\ref{sec:method:eval} (Appendix~\ref{app:eval}). The
evaluation data do not overlap the training data: no series of BEDTime, CaTS-Bench or the four
forecasting corpora appears among the training fragments (Appendix~\ref{app:overlap}).

Every trained model starts from the same Qwen2.5-VL-3B-Instruct weights and is prompted with
the same instruction. We compare \method\ against the untuned initialization; against SFT
distilled from Qwen2.5-VL-72B-Instruct captions of the same charts; against RL with a
CapRL-style answerability reward \citep{xing2025caprl}; and against RL with an LLM-as-judge
reward. The three RL variants share the policy-optimization hyperparameters and the frozen
Qwen2.5-14B-Instruct behind the reward, which serves as our verifier, as the answerer of the
answerability reward and as the judge, so they differ only in the form of the reward
(Appendix~\ref{app:baselines}). The $72$B teacher, two further off-the-shelf VLMs, three text
LLMs that read the values as digits, and a time-series language model are reported for
reference. Criterion 2 is scored by two frozen LLMs that take no part in training any model,
GLM-4-9B and Phi-4-14B, and every identification number we report is the mean of the two;
readings under three other verifiers are in Appendix~\ref{app:results:verifiers}. Criterion 1 uses a frozen DeBERTa long-NLI model, and
Criterion 3 fits a frozen EmbeddingGemma-300M with a LoRA adapter and an MLP head under an
identical recipe for every captioner, reported as mean and standard deviation over five
training seeds. Training, prompting and predictor hyperparameters are in
Appendix~\ref{app:train}.

\subsection{Main results}
\label{sec:exp:criteria12}

\begin{table}[t]
\centering
\caption{Entailment and identification on two captioning benchmarks.
\textsc{c}$\to$\textsc{s} is short for caption-to-series (\textsc{cap}$\to$\textsc{ser}) identification. \textsc{s}$\to$\textsc{c} is short for series-to-caption (\textsc{ser}$\to$\textsc{cap}) identification. Each identification cell is the mean over two frozen verifiers that took no part in training, GLM-4-9B and Phi-4-14B; random chance is $\frac{1}{K}=0.25$. CaTS-Bench entailment is scored against rewritten references (Appendix~\ref{app:cats-rewrite}). Bold marks the best of the four models fine-tuned from the same initialization.}
\label{tab:criteria12}
\small
\setlength{\tabcolsep}{4.5pt}
\begin{tabular}{@{}llcccccc@{}}
\toprule
& & \multicolumn{3}{c}{BEDTime} & \multicolumn{3}{c}{CaTS-Bench} \\
\cmidrule(lr){3-5} \cmidrule(lr){6-8}
Type & Captioner & Entailment & \textsc{c}$\to$\textsc{s} & \textsc{s}$\to$\textsc{c} &
Entailment & \textsc{c}$\to$\textsc{s} & \textsc{s}$\to$\textsc{c} \\
\midrule
\multirow{4}{*}{VLM} & Qwen2.5-VL-3B & 0.296 & 0.550 & 0.495 & 0.352 & 0.751 & 0.721 \\
  & Qwen2.5-VL-72B & 0.320 & 0.605 & 0.484 & 0.440 & 0.803 & 0.643 \\
  & Qwen3-VL-8B & 0.320 & 0.669 & 0.498 & 0.407 & 0.817 & 0.626 \\
  & InternVL3-14B & 0.357 & 0.562 & 0.512 & 0.506 & 0.766 & 0.701 \\
\midrule
\multirow{3}{*}{LLM} & Phi-3.5-mini & 0.180 & 0.909 & 0.807 & 0.318 & 0.980 & 0.890 \\
  & Qwen2.5-7B & 0.240 & 0.916 & 0.823 & 0.400 & 0.985 & 0.900 \\
  & Qwen2.5-14B & 0.297 & 0.919 & 0.838 & 0.535 & 0.991 & 0.938 \\
\midrule
TSLM & ChatTS-14B & 0.272 & 0.683 & 0.721 & 0.292 & 0.882 & 0.863 \\
\midrule
\multicolumn{8}{@{}l}{\emph{Fine-tuned from Qwen2.5-VL-3B}} \\

\multirow{4}{*}{VLM} & SFT on 72B's captions & 0.281 & 0.570 & 0.448 & 0.408 & 0.760 & 0.616 \\
  & RL, answerability & 0.286 & 0.448 & 0.443 & 0.299 & 0.599 & 0.560 \\
  & RL, LLM-as-judge & 0.118 & 0.544 & 0.412 & 0.033 & 0.694 & 0.589 \\
  & \method & \textbf{0.400} & \textbf{0.709} & \textbf{0.638} & \textbf{0.611} & \textbf{0.886} & \textbf{0.868} \\
\bottomrule
\end{tabular}
\end{table}

Table~\ref{tab:criteria12} reports the performance of all models on the two benchmarks. Among the four models fine-tuned from the same
initialization, \method\ is best in all six columns. The two RL baselines show how much the reward design matters. The answerability reward ends up below
the initialization it started from on all four identification columns. The
LLM-as-judge reward loses less identification than the answerability reward on three of the four columns, but its entailment score collapses to $0.118$ and $0.033$, far beneath the base model.
Against the 72B teacher itself, \method\ is ahead in all six columns with $1/24$ of the parameters, which shows that an appropriate RL pipeline can raise a VLM's time series captioning ability beyond what distillation offers. Another finding concerns the three
text LLMs, which read the raw values instead of the chart. Their identification scores are far above
every chart-reading model on both benchmarks, but this advantage does not carry over to entailment: even the
largest of them trails \method. Reading the raw values thus helps mainly with the fine detail that tells one series from
another, and much less with describing the overall shape and events of the time series.

\begin{table}[t]
\centering
\caption{Information bottleneck on time series tasks:
Mean and standard deviation of test MSE over five training seeds of the predictor are reported. The naive baseline is last-value persistence for forecasting and mean training window for reconstruction.
Bold marks every fine-tuned model not significantly worse than the best
in its column; significance tests are deferred to Appendix~\ref{app:sig}.}
\label{tab:readout}
\footnotesize
\setlength{\tabcolsep}{5.5pt}
\begin{tabular}{@{}lcccccccc@{}}
\toprule
& \multicolumn{4}{c}{\textbf{Forecasting}}
& \multicolumn{4}{c}{\textbf{Reconstruction}} \\
\cmidrule(lr){2-5} \cmidrule(lr){6-9}
Captioner & ETTh2 & ETTm2 & Saug. & Elec. & ETTh2 & ETTm2 & Saug. & Elec. \\
\midrule
Naive  baseline & 0.3047 & 0.0596 & 1.2955 & 0.9078 & 0.9501 & 0.8335 & 1.1748 & 0.8634 \\
Empty caption & \shortstack{0.8446\\[-2.5pt]{\tiny$\pm$0.0101}} & \shortstack{0.8001\\[-2.5pt]{\tiny$\pm$0.0042}} & \shortstack{1.0601\\[-2.5pt]{\tiny$\pm$0.0016}} & \shortstack{0.9129\\[-2.5pt]{\tiny$\pm$0.0041}} & \shortstack{0.9284\\[-2.5pt]{\tiny$\pm$0.0306}} & \shortstack{0.8274\\[-2.5pt]{\tiny$\pm$0.0029}} & \shortstack{1.1764\\[-2.5pt]{\tiny$\pm$0.0008}} & \shortstack{0.8667\\[-2.5pt]{\tiny$\pm$0.0021}} \\
Shuffled caption & \shortstack{0.8437\\[-2.5pt]{\tiny$\pm$0.0087}} & \shortstack{0.8040\\[-2.5pt]{\tiny$\pm$0.0046}} & \shortstack{1.0619\\[-2.5pt]{\tiny$\pm$0.0042}} & \shortstack{0.9124\\[-2.5pt]{\tiny$\pm$0.0043}} & \shortstack{0.9347\\[-2.5pt]{\tiny$\pm$0.0121}} & \shortstack{0.8270\\[-2.5pt]{\tiny$\pm$0.0015}} & \shortstack{1.1774\\[-2.5pt]{\tiny$\pm$0.0018}} & \shortstack{0.8663\\[-2.5pt]{\tiny$\pm$0.0010}} \\
\addlinespace[2pt]
Qwen2.5-VL-3B & \shortstack{0.3062\\[-2.5pt]{\tiny$\pm$0.0086}} & \shortstack{0.1012\\[-2.5pt]{\tiny$\pm$0.0047}} & \shortstack{0.9419\\[-2.5pt]{\tiny$\pm$0.0119}} & \shortstack{0.5751\\[-2.5pt]{\tiny$\pm$0.0351}} & \shortstack{0.2445\\[-2.5pt]{\tiny$\pm$0.0052}} & \shortstack{0.0633\\[-2.5pt]{\tiny$\pm$0.0029}} & \shortstack{0.6852\\[-2.5pt]{\tiny$\pm$0.0173}} & \shortstack{0.4352\\[-2.5pt]{\tiny$\pm$0.0156}} \\
Qwen2.5-VL-72B & \shortstack{0.2149\\[-2.5pt]{\tiny$\pm$0.0094}} & \shortstack{0.0451\\[-2.5pt]{\tiny$\pm$0.0009}} & \shortstack{0.8521\\[-2.5pt]{\tiny$\pm$0.0061}} & \shortstack{0.3089\\[-2.5pt]{\tiny$\pm$0.0094}} & \shortstack{0.1134\\[-2.5pt]{\tiny$\pm$0.0065}} & \shortstack{0.0288\\[-2.5pt]{\tiny$\pm$0.0022}} & \shortstack{0.4444\\[-2.5pt]{\tiny$\pm$0.0113}} & \shortstack{0.1993\\[-2.5pt]{\tiny$\pm$0.0077}} \\
\midrule
\multicolumn{9}{@{}l}{\emph{Fine-tuned from Qwen2.5-VL-3B}} \\
SFT on 72B's captions & \shortstack{0.2595\\[-2.5pt]{\tiny$\pm$0.0151}} & \shortstack{0.0615\\[-2.5pt]{\tiny$\pm$0.0021}} & \shortstack{0.9026\\[-2.5pt]{\tiny$\pm$0.0216}} & \shortstack{0.3480\\[-2.5pt]{\tiny$\pm$0.0086}} & \shortstack{0.1478\\[-2.5pt]{\tiny$\pm$0.0044}} & \shortstack{0.0333\\[-2.5pt]{\tiny$\pm$0.0016}} & \shortstack{0.5259\\[-2.5pt]{\tiny$\pm$0.0177}} & \shortstack{0.2144\\[-2.5pt]{\tiny$\pm$0.0084}} \\
RL, answerability & \shortstack{0.3780\\[-2.5pt]{\tiny$\pm$0.0168}} & \shortstack{0.2110\\[-2.5pt]{\tiny$\pm$0.0159}} & \shortstack{0.9384\\[-2.5pt]{\tiny$\pm$0.0104}} & \shortstack{0.3258\\[-2.5pt]{\tiny$\pm$0.0151}} & \shortstack{0.3391\\[-2.5pt]{\tiny$\pm$0.0185}} & \shortstack{0.1883\\[-2.5pt]{\tiny$\pm$0.0151}} & \shortstack{0.5309\\[-2.5pt]{\tiny$\pm$0.0091}} & \shortstack{0.2608\\[-2.5pt]{\tiny$\pm$0.0095}} \\
RL, LLM-as-judge & \shortstack{0.2384\\[-2.5pt]{\tiny$\pm$0.0134}} & \shortstack{0.0630\\[-2.5pt]{\tiny$\pm$0.0077}} & \shortstack{0.8619\\[-2.5pt]{\tiny$\pm$0.0217}} & \shortstack{0.3388\\[-2.5pt]{\tiny$\pm$0.0113}} & \shortstack{0.1373\\[-2.5pt]{\tiny$\pm$0.0104}} & \shortstack{0.0325\\[-2.5pt]{\tiny$\pm$0.0019}} & \shortstack{\textbf{0.4517}\\[-2.5pt]{\tiny$\pm$0.0027}} & \shortstack{0.2248\\[-2.5pt]{\tiny$\pm$0.0029}} \\
\method & \shortstack{\textbf{0.1806}\\[-2.5pt]{\tiny$\pm$0.0078}} & \shortstack{\textbf{0.0541}\\[-2.5pt]{\tiny$\pm$0.0022}} & \shortstack{\textbf{0.8175}\\[-2.5pt]{\tiny$\pm$0.0196}} & \shortstack{\textbf{0.2042}\\[-2.5pt]{\tiny$\pm$0.0091}} & \shortstack{\textbf{0.0989}\\[-2.5pt]{\tiny$\pm$0.0033}} & \shortstack{\textbf{0.0294}\\[-2.5pt]{\tiny$\pm$0.0010}} & \shortstack{\textbf{0.4463}\\[-2.5pt]{\tiny$\pm$0.0054}} & \shortstack{\textbf{0.1357}\\[-2.5pt]{\tiny$\pm$0.0068}} \\
\bottomrule
\end{tabular}
\end{table}

Table~\ref{tab:readout} reports the test error of a caption-only predictor on forecasting and reconstruction. The predictor is an EmbeddingGemma-300M \citep{gemma300m} equipped with a LoRA \citep{hu2021lora} adapter and a lightweight MLP decoder head, reading the
caption alone.
The three baseline rows set the scale of the errors.
\method\ has the lowest error in all eight cells, significantly so in seven, the exception
being reconstruction on Saugeen where the LLM-as-judge reward ties it. Compared to Qwen2.5-VL-72B, \method wins five cells, ties two and loses one (forecasting on ETTm2), so the $3$B model is on par with a model $24\times$ its size. Two patterns from Table~\ref{tab:criteria12} recur: the
answerability reward scores significantly worse for forecasting than the untuned initialization
on ETTh2 and ETTm2, while the LLM-as-judge reward, whose entailment collapsed there, is the
second-best captioner in five of the eight cells, so a caption can entail the reference and still be of little use
downstream.

\Needspace{4\baselineskip}
\subsection{Ablations}
\label{sec:exp:ablation}

\begin{figure}[t]
\centering
\includegraphics[width=\textwidth]{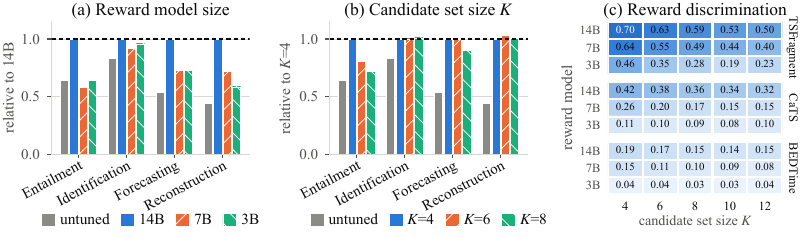}
\caption{Ablation on the reward's three knobs. (a) three reward models and (b)
$K\in\{4,6,8\}$, where $K$ is the cardinality of the candidate set in each identification
question and $K-1$ the number of distractors; the identification bar is
\textsc{ser}$\to$\textsc{cap} on CaTS-Bench under the two evaluation verifiers. (c) re-scores reference captions
of three domains, written by no captioner, under every combination of reward model size (down)
and $K$ (across); each cell is accuracy above chance, $(\mathrm{acc}-1/K)/(1-1/K)$, and
darker is higher. Every number behind the three panels is in Appendix~\ref{app:ablation}.}
\label{fig:ablation}
\end{figure}

%
%
\begin{wraptable}[13]{r}{0.46\textwidth}
\vspace{-\intextsep}
\centering
\caption{Nearest against random distractors.}
\label{tab:negrand}
\footnotesize
\renewcommand{\arraystretch}{1.0}
\setlength{\tabcolsep}{4pt}
\begin{tabular}{@{}lcc@{}}
\toprule
& Nearest & Random \\
Distractors & (\textbf{ours}) & \\
\midrule
Reward acc.\ over chance & 0.872 & 0.982 \\
\midrule
Entailment & \textbf{0.506} & 0.354 \\
Ident.\ \textsc{c}$\to$\textsc{s} & \textbf{0.797} & 0.704 \\
Ident.\ \textsc{s}$\to$\textsc{c} & \textbf{0.753} & 0.744 \\
Caption length (chars) & 667 & 738 \\
Forecasting rel.\ MSE & \textbf{1.000} & 1.212 \\
Reconstruction rel.\ MSE & \textbf{1.000} & 1.142 \\
\bottomrule
\end{tabular}
\end{wraptable}

In the ablation study, we investigate three factors that affect the performance of \method: the size of the reward model, the cardinality of the candidate set, and the selection of distractors.
Figure~\ref{fig:ablation}(a) shows that shrinking the reward model
hurts all four criteria. With reward models smaller than 14B, the fine-tuned model
does not improve entailment over the untuned model (the 3B verifier ties it), which indicates that our reward still requires a certain degree of time series comprehension from the LLM: less than writing questions or grading a caption, but more than a 7B model has.
Figure~\ref{fig:ablation}(b) shows that more distractors do not help in general: entailment drops markedly, identification and reconstruction tie, and forecasting error rises by $11\%$ at $K=8$. Figure~\ref{fig:ablation}(c) explains both observations by examining the reward signal under each setting on reference captions that no captioner wrote. Caption-to-series identification accuracy falls with a smaller reward model and with more distractors, which makes the reward signal noisier; the 7B and 3B verifiers are close to chance on CaTS-Bench and BEDTime even on these captions, so the drop in (a) comes from the verifier rather than from the captions it graded.

Another important factor in \method\ is negative selection, the rule that draws the distractors.
\method\ draws the $K-1$
distractors of a training item from its nearest neighbors in a four-statistic feature
space; Table~\ref{tab:negrand} replaces that rule
with a uniform random draw from time series of the same length and compares the two fine-tuned models.
Its caption rows average BEDTime and CaTS-Bench, and its forecasting and reconstruction rows are geometric means
over the four corpora of the error relative to \method; Appendix~\ref{app:ablation} gives
every cell.
Hard negatives give the better captioner on every criterion.
The random-distractor variant in fact scores higher on its own training signal, $0.982$ above
chance against $0.872$ for \method, but that only means the question got easier. A random distractor differs from the true series in level and range, so
stating the approximate level is enough to earn the reward, and the caption need not say more. Nearest neighbors agree on those coarse statistics, so the caption has to describe the
finer shape before the verifier can tell them apart.

\subsection{Case study}
\label{sec:exp:case}

\makeatletter
\@ifundefined{textcolor}{%
  \providecommand{\cok}[1]{\textbf{#1}}%
  \providecommand{\cpos}[1]{#1}%
  \providecommand{\cbad}[1]{\textbf{\itshape #1}}%
  \providecommand{\cnot}[1]{\textit{#1}}%
}{%
  \definecolor{okc}{rgb}{0.05,0.43,0.26}%
  \definecolor{posc}{rgb}{0.12,0.37,0.66}%
  \definecolor{badc}{rgb}{0.78,0.16,0.12}%
  \definecolor{filc}{rgb}{0.55,0.55,0.55}%
  \providecommand{\cok}[1]{\textcolor{okc}{#1}}%
  \providecommand{\cpos}[1]{\textcolor{posc}{#1}}%
  \providecommand{\cbad}[1]{\textcolor{badc}{\textbf{#1}}}%
  \providecommand{\cnot}[1]{\textcolor{filc}{\textit{#1}}}%
}
\@ifundefined{uline}{\providecommand{\ev}[1]{\underline{#1}}}%
                    {\providecommand{\ev}[1]{\uline{#1}}}
\providecommand{\elip}{\textup{[\dots]}}
\makeatother

\begin{figure}[t]
\centering
{\scriptsize
\setlength{\tabcolsep}{0pt}
\begin{tabular}{@{}>{\sloppy\raggedright\arraybackslash}p{0.585\linewidth}@{\hspace{0.04\linewidth}}>{\sloppy\raggedright\arraybackslash}p{0.375\linewidth}@{}}
\toprule
\vspace{0pt}\includegraphics[width=\linewidth]{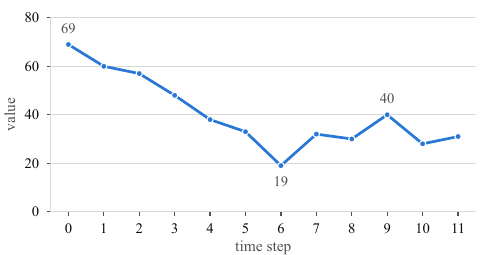}
&
\textbf{\method\ (ours).} {\tiny $668$ characters, $9$ time positions, $9$
values.} \newline
\cok{The time series \ev{starts with an initial value of approximately 68}. It then shows a
\ev{gradual decline, maintaining a value around 60 by time step 2}. Following this, there's a
steady decrease, with a value around 58 by time step 3. The value continues to drop, reaching
approximately 42 by time step 5. There's a \ev{notable dip, dropping to about 16 by time step
6}. It then experiences a \ev{slight recovery, maintaining a value around 31 by time steps 7 and
8}. There's a subsequent upward movement,} \cbad{peaking at a value of about 33 by time step 9.}
\cok{Finally, the series settles around a value of approximately 30 by time step 11, with a final
value of about 30 at the 11th time step.} \\
\midrule
\textbf{RL, LLM-as-judge reward.} {\tiny $1{,}424$ characters, $9$ time positions, $10$ values.}
\newline
\cnot{The line chart depicts a single time series with a y-axis labeled ``value'' and an x-axis
labeled ``time''.} \cbad{The y-axis ranges approximately from 0 to 70 and marks increments of
10,} \cnot{while the x-axis ranges approximately from 0 to 11 and marks integer values. The plot
shows a blue line with circular blue markers at each data point.} \cbad{The line \ev{starts
slightly above 60} on the y-axis at the beginning of the x-axis,} \cok{\ev{descends to just below
60 around the 2nd point} on the x-axis, drops to about 55 around the 3rd point, continues to
about 45 around the 4th point, drops to around 32 around the 5th point, \ev{dips to about 15
around the 6th point}, \ev{rises to about 30 around the 7th point},} \cbad{peaks at about 35
around the 8th point,} \cok{and then slightly decreases to about 30 around the 10th point, ending
just below 30 at the 11th point on the x-axis.} \cnot{This description captures every detail of
the chart, including the exact values at each point, the exact position on the axes, the exact
trend of the line, and the exact position of the markers. Anyone reading this description would
be able to recreate this specific line chart exactly, as all details are precise and complete.
The sequence and magnitude of the points, their exact values, the exact position on the axes, and
the exact trend of the line are all exactly as described. There are no nuances or variations;
every aspect of the chart is accounted for in this detailed description.}
&
\textbf{RL, answerability reward.} {\tiny $918$ characters, $21$ time positions, $0$ values.}
\newline
\cnot{The line chart depicts a single time series dataset. Here is a detailed description of the
series: 1. Key Points and Values:} \cpos{The \ev{highest values are at time 0}. The
\ev{lowest values are at time 5-6}.} \cnot{2. Key Features:} \cpos{The values \ev{decrease
rapidly from time 0 to time 3}. There is a steep decline from time 2 to time 4. The values remain
relatively stable from time 4 to time 5.} \cnot{3. Notable Features:} \cpos{The values drop
sharply from time 0 to time 1.} \cbad{The values drop sharply from time 1 to time 2.}
\cpos{The values drop sharply from time 2 to time 3.} \cnot{4. Key Points:} \cpos{Maximum values:
Around time 0. Minimum values: Around time 6.} \cnot{5. Key Points of Interest:} \cpos{The
highest values are between time 0 and time 1. The lowest values are between time 5 and time 6.
This series is characterized by a steep decline followed by \ev{a slight rise and then a steady
level}.} \\
\bottomrule
\end{tabular}}
\caption{Case study on captions from three RL-fine-tuned models for the series at the
top left. Spans are classed as
\cok{green}, a fact pinned to a time and a magnitude the series supports (within $0.35\sigma$ of
the series over the interval it points at); \cpos{blue}, a position with no magnitude;
\cbad{red}, a claim the series contradicts; \cnot{gray}, content irrelevant to the series.
\ev{Underline} marks a span covering one of the series' four events (opening maximum, fall,
trough at $t{=}6$, rebound), which all three captions visit in order.}
\label{fig:case}
\end{figure}

Figure~\ref{fig:case} shows the captions that the three RL reward designs produce for the same series, and it
makes the weakness of both baseline rewards concrete. The answerability reward is limited by the
model that sets its questions: posing and answering questions about a time series is far harder
for an LLM or a VLM than doing so about a natural image. Only $28\%$ of the questions the VLM wrote survive a check against the raw values with the answer the VLM gave (Appendix~\ref{app:baselines}), and most survivors merely ask which interval holds an extremum, so the captions this reward trains do not
describe a time series well. The LLM-as-judge reward behaves as reported for image
captioning, where an LLM judge is easy to hack: much of the caption is text about the caption itself, claiming that it is precise and
complete, rather than about the time series. \method, in contrast, describes the events of the series in order and gives the key values,
and its caption is the shortest of the three.

\section{Related work}
\label{sec:related}

\textbf{Reinforcement learning with verifiable rewards.} RLVR replaces the learned preference model of RLHF \citep{ouyang2022training} with a
programmatic check on the sampled output, optimized by critic-free group estimators such as GRPO
\citep{shao2024deepseekmath} and RLOO \citep{kool2019buy,ahmadian2024back}. RLVR applies most directly where a
rule decides correctness outright \citep{shao2024deepseekmath,guo2025deepseekr1,zhang2026chart}, and has since been extended to open-ended tasks by designing verifiers for them. Recent reward designs include rubrics decomposing a subjective answer into checkable criteria
\citep{gunjal2026rubrics,huang2025reinforcement}, outcome checks against an external environment \citep{jin2025searchr1,zheng2025deepresearcher}, and downstream utility \citep{berthon2025language}.

\textbf{Time series captioning.} Early systems verbalize an explicit model of the data, from Gaussian-process
reports to utility-ranked trends and executable pattern programs
\citep{lloyd2014automatic,jandaghi2020human,jhamtani2021truth}. Language models now write the descriptions themselves, either as dedicated
captioners or as one capability of a time-series LLM
\citep{trabelsi2025tslm,lin2024tessa,xie2024chatts,feng2026decoupling}. Reinforcement learning has been applied to the time series domain where the task already supplies a verifiable target \citep{zhang2025timemaster,fan2026timesrl,liu2026posttime,houwords}.
In our setting the description itself is the output, it is rewarded by whether a frozen verifier can identify the series from it alone, and no downstream task or label is involved.
BEDTime \citep{sen2025bedtime} and CaTS-Bench \citep{zhou2026catsbench} benchmark how well current
models caption a time series, and feature-level probes
\citep{fons2024evaluating,merrill2024language} test the same ability indirectly.

\textbf{Reinforcement learning for image captioning.}
Image captioning is closest to ours. Metric rewards score a sampled
caption against human references and became standard with the self-critical baseline
\citep{rennie2017self,liu2017improved}. Human-feedback rewards replace the metric with ratings
or written corrections \citep{ling2017teaching,seo2020reinforcing}. Model-scored rewards employ a frozen cross-modal encoder to score the caption by its similarity to the image \citep{cho2022fine,dzabraev2024vlrm} or by retrieval \citep{luo2018discriminability,dessi2023cross,chaffin2024distinctive}. The retrieval line is the closest to ours in spirit: a listener must pick the described image among distractors. It relies on a retrieval model trained on paired image-text data, which does not exist for time series, and scores against distractors drawn at random or by image similarity. \method\ replaces the listener with a frozen text LLM that reads the candidates as raw values, so no retrieval model is trained, draws the distractors from real series matched on summary statistics, and optimizes a VLM policy with verifiable rewards. Judge rewards employ an LLM judge with rubrics or verified atomic claims
\citep{huang2026rubicap,li2026claimdiff,ye2026balcaprl,tang2026cccaption}. 
Utility rewards
make the caption serve a downstream task, e.g., question answering \citep{fisch2020capwap,xing2025caprl} or safety
decision \citep{ma2026safecap}.

\section{Conclusion}
\label{sec:conclusion}

We present \method, a reinforcement learning framework that scores a time series caption by
whether the series can be identified from it. By rewarding a caption for letting a frozen text-only verifier pick its series out of
a set of close alternatives, \method\ obtains its training signal from the data alone, without reference captions or a
learned reward model. Across two
captioning benchmarks and a caption-only forecasting and reconstruction evaluation, a $3$B
captioner trained this way surpasses SFT distillation, alternative RL rewards, and the
$72$B model it could have been distilled from. Ablation studies validate the three choices
the reward adds, the scale of the verifier, the number of candidates and the rule that draws
the distractors, and show
that the hard negatives give the better captioner on every criterion. Caption-to-series identification
against close alternatives is thus a workable verifiable reward for
continuous signals, where reference captions are scarce and LLM judges are easy to hack. Longer and
multivariate series and captioners beyond VLMs are left
to future work.

\subsection*{AI use statement}

In this work, we used generative AI tools to generate teacher captions for the
supervised fine-tuning (SFT) baseline, to write and verify the multiple-choice questions of
the answerability baseline (Appendix~\ref{app:baselines}), and to rewrite the CaTS-Bench
reference captions used by Criterion 1 (Claude Opus 5, Appendix~\ref{app:cats-rewrite}); the
rewritten references were checked by a script and reviewed by the authors. We did
not use generative AI tools to implement the proposed algorithm. Additionally, we used AI tools to organize our
code, prepare figures and tables, check
the manuscript for grammatical and spelling errors, and polish language. The authors are responsible
for the final manuscript, code, and other research artifacts, including
AI-assisted content.

\subsection*{Ethics statement}

This work studies automated time series captioning using existing datasets and
pretrained models. Generated captions may omit relevant events or misstate trends
and numerical values. The identification reward measures whether a verifier can
distinguish the target series from alternatives; it does not guarantee the
correctness of every statement in a caption. Captions should therefore be checked
against the underlying series before informing consequential decisions.
Applications involving sensitive time series require appropriate privacy
protections and evaluation in the intended domain.

\subsection*{Reproducibility statement}
The reward and the distractor construction are defined in \S\ref{sec:method:ours}, and the
three evaluation criteria in \S\ref{sec:method:eval}. Appendix~\ref{app:train} gives the
implementation: the training data, policy-optimization hyperparameters and reward configuration
of \method\ ($K$, $\lambda$, $\eta$, the verifier and its decoding mode), the data, reward and
checkpoint-selection rule of every baseline, and what each ablation setting changes.
Appendix~\ref{app:eval} gives the evaluation: the benchmark subsets and their sizes, the
candidate-set construction, the rewriting of the CaTS-Bench references, the NLI model and
threshold, the windowing of the forecasting corpora and the recipe of the caption-only
predictor, which is held fixed across captioners, the significance tests behind the bold
entries of Table~\ref{tab:readout}, and every cell of that table with its across-seed spread and
the seeds used. Appendix~\ref{app:ablation} reports every captioner under five verifiers and the
numbers behind every ablation panel. Appendix~\ref{app:prompts} gives every prompt verbatim. The rendering, training,
reward-server and evaluation code, every prompt, the identifiers of the $10{,}000$ training
fragments and the rewritten CaTS-Bench references are available at
\url{https://github.com/HaochenZhang717/LineupRL-official-repo}; the rendered charts are
regenerated deterministically by the released renderer. The captions written by every captioner
in Tables~\ref{tab:criteria12} and~\ref{tab:readout} and the scripts that regenerate every table
and figure from the raw metric files will be released with the paper.

% ...

\bibliographystyle{style/icml2025}
\bibliography{999_reference}

@inproceedings{rennie2017self,
  title={Self-critical sequence training for image captioning},
  author={Rennie, Steven J and Marcheret, Etienne and Mroueh, Youssef and Ross, Jerret and Goel, Vaibhava},
  booktitle={2017 IEEE conference on computer vision and pattern recognition (CVPR)},
  pages={1179--1195},
  year={2017},
  organization={IEEE}
}

@inproceedings{houwords,
  title={Words Towards Explainability: Caption Label-Free Learning via Dual Loop Agentic Time Series Captioning},
  author={Hou, Difei and Yue, Jiaqi and Zhao, Chunhui},
  booktitle={Forty-third International Conference on Machine Learning},
  year={2026}
}

@article{yu2007choosing,
  title={Choosing the content of textual summaries of large time-series data sets},
  author={Yu, Jin and Reiter, Ehud and Hunter, Jim and Mellish, Chris},
  journal={Natural Language Engineering},
  volume={13},
  number={1},
  pages={25--49},
  year={2007},
  publisher={Cambridge University Press}
}

@article{reiter2005choosing,
  title={Choosing words in computer-generated weather forecasts},
  author={Reiter, Ehud and Sripada, Somayajulu and Hunter, Jim and Yu, Jin and Davy, Ian},
  journal={Artificial Intelligence},
  volume={167},
  number={1-2},
  pages={137--169},
  year={2005},
  publisher={Elsevier}
}

@article{chu2025sft,
  title={Sft memorizes, rl generalizes: A comparative study of foundation model post-training},
  author={Chu, Tianzhe and Zhai, Yuexiang and Yang, Jihan and Tong, Shengbang and Xie, Saining and Schuurmans, Dale and Le, Quoc V and Levine, Sergey and Ma, Yi},
  journal={arXiv preprint arXiv:2501.17161},
  year={2025}
}

@inproceedings{shenfeld2025razor,
  title={Rl's razor: Why online reinforcement learning forgets less},
  author={Shenfeld, Idan and Pari, Jyothish and Agrawal, Pulkit},
  booktitle={International Conference on Learning Representations},
  volume={2026},
  pages={59839--59864},
  year={2026}
}

@article{lambert2024tulu,
  title={Tulu 3: Pushing frontiers in open language model post-training},
  author={Lambert, Nathan and Morrison, Jacob and Pyatkin, Valentina and Huang, Shengyi and Ivison, Hamish and Brahman, Faeze and Miranda, Lester James V and Liu, Alisa and Dziri, Nouha and Lyu, Shane and others},
  journal={arXiv preprint arXiv:2411.15124},
  year={2024}
}

@article{portet2009automatic,
  title={Automatic generation of textual summaries from neonatal intensive care data},
  author={Portet, Fran{\c{c}}ois and Reiter, Ehud and Gatt, Albert and Hunter, Jim and Sripada, Somayajulu and Freer, Yvonne and Sykes, Cindy},
  journal={Artificial Intelligence},
  volume={173},
  number={7-8},
  pages={789--816},
  year={2009},
  publisher={Elsevier}
}

@inproceedings{kukich1983design,
  title={Design of a knowledge-based report generator},
  author={Kukich, Karen},
  booktitle={21st Annual Meeting of the Association for Computational Linguistics},
  pages={145--150},
  year={1983}
}

@article{law2005comparison,
  title={A comparison of graphical and textual presentations of time series data to support medical decision making in the neonatal intensive care unit},
  author={Law, Anna S and Freer, Yvonne and Hunter, Jim and Logie, Robert H and McIntosh, Neil and Quinn, John},
  journal={Journal of clinical monitoring and computing},
  volume={19},
  number={3},
  pages={183--194},
  year={2005},
  publisher={Springer}
}

@article{hu2021lora,
  title={Lora: Low-rank adaptation of large language models},
  author={Hu, Edward J and Shen, Yelong and Wallis, Phillip and Allen-Zhu, Zeyuan and Li, Yuanzhi and Wang, Shean and Wang, Lu and Chen, Weizhu},
  journal={arXiv preprint arXiv:2106.09685},
  year={2021}
}

@article{gemma300m,
  title={Embeddinggemma: Powerful and lightweight text representations},
  author={Vera, Henrique Schechter and Dua, Sahil and Zhang, Biao and Salz, Daniel and Mullins, Ryan and Panyam, Sindhu Raghuram and Smoot, Sara and Naim, Iftekhar and Zou, Joe and Chen, Feiyang and others},
  journal={arXiv preprint arXiv:2509.20354},
  year={2025}
}

@article{ge2025t2s,
  title={T2s: High-resolution time series generation with text-to-series diffusion models},
  author={Ge, Yunfeng and Li, Jiawei and Zhao, Yiji and Wen, Haomin and Li, Zhao and Qiu, Meikang and Li, Hongyan and Jin, Ming and Pan, Shirui},
  journal={arXiv preprint arXiv:2505.02417},
  year={2025}
}

@inproceedings{luo2018discriminability,
  title={Discriminability objective for training descriptive captions},
  author={Luo, Ruotian and Price, Brian and Cohen, Scott and Shakhnarovich, Gregory},
  booktitle={Proceedings of the IEEE conference on computer vision and pattern recognition},
  pages={6964--6974},
  year={2018}
}

@inproceedings{liu2017improved,
  title={Improved image captioning via policy gradient optimization of spider},
  author={Liu, Siqi and Zhu, Zhenhai and Ye, Ning and Guadarrama, Sergio and Murphy, Kevin},
  booktitle={2017 IEEE International Conference on Computer Vision (ICCV)},
  pages={873--881},
  year={2017},
  organization={IEEE}
}

@article{dzabraev2024vlrm,
  title={Vlrm: Vision-language models act as reward models for image captioning},
  author={Dzabraev, Maksim and Kunitsyn, Alexander and Ivaniuta, Andrei},
  journal={arXiv preprint arXiv:2404.01911},
  year={2024}
}

@inproceedings{chaffin2024distinctive,
  title={Distinctive image captioning: Leveraging ground truth captions in clip guided reinforcement learning},
  author={Chaffin, Antoine and Kijak, Ewa and Claveau, Vincent},
  booktitle={2024 IEEE International Conference on Image Processing (ICIP)},
  pages={2550--2556},
  year={2024},
  organization={IEEE}
}

@article{huang2026rubicap,
  title={RubiCap: Rubric-Guided Reinforcement Learning for Dense Image Captioning},
  author={Huang, Tzu-Heng and Salekin, Sirajul and Movellan, Javier and Sala, Frederic and Bilkhu, Manjot},
  journal={arXiv preprint arXiv:2603.09160},
  year={2026}
}

@inproceedings{cho2022fine,
  title={Fine-grained image captioning with clip reward},
  author={Cho, Jaemin and Yoon, Seunghyun and Kale, Ajinkya and Dernoncourt, Franck and Bui, Trung and Bansal, Mohit},
  booktitle={Findings of the Association for Computational Linguistics: NAACL 2022},
  pages={517--527},
  year={2022}
}

@article{ye2026balcaprl,
  title={BalCapRL: A Balanced Framework for RL-Based MLLM Image Captioning},
  author={Ye, Shaokai and Saveris, Vasileios and Qian, Yihao and Hu, Jiaming and Amirloo, Elmira and Grasch, Peter},
  journal={arXiv preprint arXiv:2605.07394},
  year={2026}
}

@inproceedings{xing2025caprl,
  title={Caprl: Stimulating dense image caption capabilities via reinforcement learning},
  author={Xing, Long and Dong, Xiaoyi and Zang, Yuhang and Cao, Yuhang and Liang, Jianze and Huang, Qidong and Wang, Jiaqi and Wu, Feng and Lin, Dahua},
  booktitle={International Conference on Learning Representations},
  volume={2026},
  pages={13066--13093},
  year={2026}
}

@inproceedings{fisch2020capwap,
  title={CapWAP: Image captioning with a purpose},
  author={Fisch, Adam and Lee, Kenton and Chang, Ming-Wei and Clark, Jonathan H and Barzilay, Regina},
  booktitle={Proceedings of the 2020 Conference on Empirical Methods in Natural Language Processing (EMNLP)},
  pages={8755--8768},
  year={2020}
}

@article{ma2026safecap,
  title={SafeCap: Improving LVLM Safety with Image Captioning Reinforcement Learning},
  author={Ma, Caoyuan and Liu, Wenpu and Xie, Weichu and Gu, Tian and Zhao, Shilei and Min, Lingxi and Dong, Shuai and Xu, Yuqi and Zhao, Ji and Wang, Ziyue and others},
  journal={arXiv preprint arXiv:2608.10513},
  year={2026}
}

@article{tang2026cccaption,
  title={CCCaption: Dual-Reward Reinforcement Learning for Complete and Correct Image Captioning},
  author={Tang, Zhijiang and Wang, Linhua and Qi, Jiaxin and Jiang, Weihao and Hou, Peng and Zeng, Anxiang and Huang, Jianqiang},
  journal={arXiv preprint arXiv:2602.21655},
  year={2026}
}

@inproceedings{dessi2023cross,
  title={Cross-domain image captioning with discriminative finetuning},
  author={Dess{\`\i}, Roberto and Bevilacqua, Michele and Gualdoni, Eleonora and Rakotonirina, Nathana{\"e}l Carraz and Franzon, Francesca and Baroni, Marco},
  booktitle={2023 IEEE/CVF Conference on Computer Vision and Pattern Recognition (CVPR)},
  pages={6935--6944},
  year={2023},
  organization={IEEE}
}

@article{li2026claimdiff,
  title={ClaimDiff-RL: Fine-Grained Caption Reinforcement Learning through Visual Claim Comparison},
  author={Li, Tianle and Shen, Xuyang and Ma, Yan and Guo, Rongxin and Chen, Shaoxiang and Chen, Jiacheng and Wang, Haochen and Tang, Hongyang and Zhou, Yucong and Cheng, Yu},
  journal={arXiv preprint arXiv:2605.20278},
  year={2026}
}

@inproceedings{seo2020reinforcing,
  title={Reinforcing an image caption generator using off-line human feedback},
  author={Seo, Paul Hongsuck and Sharma, Piyush and Levinboim, Tomer and Han, Bohyung and Soricut, Radu},
  booktitle={Proceedings of the AAAI conference on artificial intelligence},
  volume={34},
  number={03},
  pages={2693--2700},
  year={2020}
}

@article{ling2017teaching,
  title={Teaching machines to describe images with natural language feedback},
  author={Fidler, Sanja and others},
  journal={Advances in Neural Information Processing Systems},
  volume={30},
  year={2017}
}

@inproceedings{fons2024evaluating,
  title={Evaluating large language models on time series feature understanding: A comprehensive taxonomy and benchmark},
  author={Fons, Elizabeth and Kaur, Rachneet and Palande, Soham and Zeng, Zhen and Balch, Tucker and Veloso, Manuela and Vyetrenko, Svitlana},
  booktitle={Proceedings of the 2024 Conference on Empirical Methods in Natural Language Processing},
  pages={21598--21634},
  year={2024}
}

@inproceedings{merrill2024language,
    title = "Language Models Still Struggle to Zero-shot Reason about Time Series",
    author = "Merrill, Mike A  and
      Tan, Mingtian  and
      Gupta, Vinayak  and
      Hartvigsen, Thomas  and
      Althoff, Tim",
    editor = "Al-Onaizan, Yaser  and
      Bansal, Mohit  and
      Chen, Yun-Nung",
    booktitle = "Findings of the Association for Computational Linguistics: EMNLP 2024",
    month = nov,
    year = "2024",
    address = "Miami, Florida, USA",
    publisher = "Association for Computational Linguistics",
    url = "https://aclanthology.org/2024.findings-emnlp.201/",
    doi = "10.18653/v1/2024.findings-emnlp.201",
    pages = "3512--3533"
}

@inproceedings{zhou2026catsbench,
    title = "{C}a{TS}-Bench: Can Language Models Describe Time Series?",
    author = "Zhou, Luca  and
      Yashwante, Pratham  and
      Fisher, Marshall  and
      Sampieri, Alessio  and
      Zhou, Zihao  and
      Galasso, Fabio  and
      Yu, Rose",
    editor = "Liakata, Maria  and
      Moreira, Viviane P.  and
      Zhang, Jiajun  and
      Jurgens, David",
    booktitle = "Findings of the {A}ssociation for {C}omputational {L}inguistics: {ACL} 2026",
    month = jul,
    year = "2026",
    address = "San Diego, California, United States",
    publisher = "Association for Computational Linguistics",
    url = "https://aclanthology.org/2026.findings-acl.1722/",
    doi = "10.18653/v1/2026.findings-acl.1722",
    pages = "34479--34519",
    ISBN = "979-8-89176-395-1"
}

@article{guo2025deepseekr1,
  title={DeepSeek-R1 incentivizes reasoning in LLMs through reinforcement learning},
  author={Guo, Daya and Yang, Dejian and Zhang, Haowei and Song, Junxiao and Wang, Peiyi and Zhu, Qihao and Xu, Runxin and Zhang, Ruoyu and Ma, Shirong and Bi, Xiao and others},
  journal={Nature},
  volume={645},
  number={8081},
  pages={633--638},
  year={2025},
  publisher={Nature Publishing Group UK London}
}

@article{shao2024deepseekmath,
  title={Deepseekmath: Pushing the limits of mathematical reasoning in open language models},
  author={Shao, Zhihong and Wang, Peiyi and Zhu, Qihao and Xu, Runxin and Song, Junxiao and Bi, Xiao and Zhang, Haowei and Zhang, Mingchuan and Li, YK and Wu, Yang and others},
  journal={arXiv preprint arXiv:2402.03300},
  year={2024}
}

@article{kool2019buy,
  title={Buy 4 reinforce samples, get a baseline for free!},
  author={Kool, Wouter and Van Hoof, Herke and Welling, Max},
  year={2019}
}

@inproceedings{ahmadian2024back,
  title={Back to basics: Revisiting REINFORCE-style optimization for learning from human feedback in LLMs},
  author={Ahmadian, Arash and Cremer, Chris and Gall{\'e}, Matthias and Fadaee, Marzieh and Kreutzer, Julia and Pietquin, Olivier and {\"U}st{\"u}n, Ahmet and Hooker, Sara},
  booktitle={Proceedings of the 62nd Annual Meeting of the Association for Computational Linguistics (Volume 1: Long Papers)},
  pages={12248--12267},
  year={2024}
}

@inproceedings{zhang2026chart,
  title={Chart-rl: Generalized chart comprehension via reinforcement learning with verifiable rewards},
  author={Zhang, Xin and Li, Xingyu and Wang, Rongguang and Miao, Ruizhong and Wang, Zheng and Wang, Yuying and Roth, Dan and Li, Chenyang},
  booktitle={Proceedings of the First Workshop on Structured Understanding, Retrieval, and Generation in the LLM Era (SURGeLLM 2026)},
  pages={107--118},
  year={2026}
}

@inproceedings{gunjal2026rubrics,
  title={Rubrics as rewards: Reinforcement learning beyond verifiable domains},
  author={Gunjal, Anisha and Wang, Anthony and Lau, Elaine and Nath, Vaskar and He, Yunzhong and Liu, Bing and Hendryx, Sean},
  booktitle={International Conference on Learning Representations},
  volume={2026},
  pages={127924--127945},
  year={2026}
}

@article{huang2025reinforcement,
  title={Reinforcement learning with rubric anchors},
  author={Huang, Zenan and Zhuang, Yihong and Lu, Guoshan and Qin, Zeyu and Xu, Haokai and Zhao, Tianyu and Peng, Ru and Hu, Jiaqi and Shen, Zhanming and Hu, Xiaomeng and others},
  journal={arXiv preprint arXiv:2508.12790},
  year={2025}
}

@article{jin2025searchr1,
  title={Search-r1: Training llms to reason and leverage search engines with reinforcement learning},
  author={Jin, Bowen and Zeng, Hansi and Yue, Zhenrui and Yoon, Jinsung and Arik, Sercan and Wang, Dong and Zamani, Hamed and Han, Jiawei},
  journal={arXiv preprint arXiv:2503.09516},
  year={2025}
}

@article{berthon2025language,
  title={Language Bottleneck Models for Qualitative Knowledge State Modeling},
  author={Berthon, Antonin and van der Schaar, Mihaela},
  journal={arXiv preprint arXiv:2506.16982},
  year={2025}
}

@inproceedings{lloyd2014automatic,
  title={Automatic construction and natural-language description of nonparametric regression models},
  author={Lloyd, James and Duvenaud, David and Grosse, Roger and Tenenbaum, Joshua and Ghahramani, Zoubin},
  booktitle={Proceedings of the AAAI Conference on Artificial Intelligence},
  volume={28},
  number={1},
  year={2014}
}

@article{jandaghi2020human,
  title={Human-like Time Series Summaries via Trend Utility Estimation},
  author={Jandaghi, Pegah and Pujara, Jay},
  journal={arXiv preprint arXiv:2001.05665},
  year={2020}
}

@inproceedings{lin2024tessa,
  title={Decoding time series with llms: A multi-agent framework for cross-domain annotation},
  author={Lin, Minhua and Chen, Zhengzhang and Liu, Yanchi and Zhao, Xujiang and Wu, Zongyu and Wang, Junxiang and Zhang, Xiang and Wang, Suhang and Chen, Haifeng},
  booktitle={Findings of the Association for Computational Linguistics: EACL 2026},
  pages={6244--6281},
  year={2026}
}

@article{feng2026decoupling,
  title={Decoupling Perception from Description: Computation-Grounded Representation Alignment between Multivariate Time Series and Language},
  author={Feng, Xinran and Xie, Yi and Zhang, Chao and Li, Ruikun and Ling, Wanyun and Li, Ziyue and Liu, Chenxi},
  journal={arXiv preprint arXiv:2608.05238},
  year={2026}
}

@article{trabelsi2025tslm,
  title={Time series language model for descriptive caption generation},
  author={Trabelsi, Mohamed and Boyd, Aidan and Cao, Jin and Uzunalioglu, Huseyin},
  journal={Engineering Applications of Artificial Intelligence},
  volume={162},
  pages={112673},
  year={2025},
  publisher={Elsevier}
}

@article{ouyang2022training,
  title={Training language models to follow instructions with human feedback},
  author={Ouyang, Long and Wu, Jeffrey and Jiang, Xu and Almeida, Diogo and Wainwright, Carroll and Mishkin, Pamela and Zhang, Chong and Agarwal, Sandhini and Slama, Katarina and Ray, Alex and others},
  journal={Advances in neural information processing systems},
  volume={35},
  pages={27730--27744},
  year={2022}
}

@inproceedings{zheng2025deepresearcher,
  title={Deepresearcher: Scaling deep research via reinforcement learning in real-world environments},
  author={Zheng, Yuxiang and Fu, Dayuan and Hu, Xiangkun and Cai, Xiaojie and Ye, Lyumanshan and Lu, Pengrui and Liu, Pengfei},
  booktitle={Proceedings of the 2025 Conference on Empirical Methods in Natural Language Processing},
  pages={414--431},
  year={2025}
}

@article{zhang2025timemaster,
  title={Timemaster: Training time-series multimodal llms to reason via reinforcement learning},
  author={Zhang, Junru and Feng, Lang and Guo, Xu and Wu, Yuhan and Dong, Yabo and Xu, Duanqing},
  journal={arXiv preprint arXiv:2506.13705},
  year={2025}
}

@article{fan2026timesrl,
  title={TimeSRL: Generalizable Time-Series Behavioral Modeling via Semantic RL-Tuned LLMs--A Case Study in Mental Health},
  author={Fan, Yuang and Xu, Lilin and Wu, Millie and Nie, Jingping and Chen, Qingyu and Yang, Yuzhe and Zhang, Zhuo and Liu, Xin and Nepal, Subigya and Jiang, Xiaofan and others},
  journal={arXiv preprint arXiv:2605.21295},
  year={2026}
}

@article{liu2026posttime,
  title={Rethinking Post-Training Recipes for Multimodal Time-Series Forecasting},
  author={Liu, Haoxin and Zhou, Yichen and Sen, Rajat and Prakash, B Aditya and Das, Abhimanyu},
  journal={arXiv preprint arXiv:2605.29401},
  year={2026}
}

@inproceedings{jhamtani2021truth,
  title={Truth-conditional captions for time series data},
  author={Jhamtani, Harsh and Berg-Kirkpatrick, Taylor},
  booktitle={Proceedings of the 2021 Conference on Empirical Methods in Natural Language Processing},
  pages={719--733},
  year={2021}
}

@inproceedings{fons2024tadacap,
  title={Tadacap: Time-series adaptive domain-aware captioning},
  author={Fons, Elizabeth and Kaur, Rachneet and Zeng, Zhen and Palande, Soham and Balch, Tucker and Vyetrenko, Svitlana and Veloso, Manuela},
  booktitle={Proceedings of the 5th ACM International Conference on AI in Finance},
  pages={54--62},
  year={2024}
}

@article{xie2024chatts,
  title={Chatts: Aligning time series with llms via synthetic data for enhanced understanding and reasoning},
  author={Xie, Zhe and Li, Zeyan and He, Xiao and Xu, Longlong and Wen, Xidao and Zhang, Tieying and Chen, Jianjun and Shi, Rui and Pei, Dan},
  journal={arXiv preprint arXiv:2412.03104},
  year={2024}
}

@article{sen2025bedtime,
  title={Bedtime: A unified benchmark for automatically describing time series},
  author={Sen, Medhasweta and Gottesman, Zachary and Qiu, Jiaxing and Bruss, C Bayan and Nguyen, Nam and Hartvigsen, Tom},
  journal={arXiv preprint arXiv:2509.05215},
  year={2025}
}

@article{aksu2024xforecast,
  title={Xforecast: Evaluating natural language explanations for time series forecasting},
  author={Aksu, Taha and Liu, Chenghao and Saha, Amrita and Tan, Sarah and Xiong, Caiming and Sahoo, Doyen},
  journal={arXiv preprint arXiv:2410.14180},
  year={2024}
}

@article{ji2026portable,
  title={Can we generate portable representations for clinical time series data using LLMs?},
  author={Ji, Zongliang and Sun, Yifei and Amaral, Andre and Goldenberg, Anna and Krishnan, Rahul G},
  journal={arXiv preprint arXiv:2603.23987},
  year={2026}
}

\clearpage
\appendix
\section{Implementation details}
\label{app:train}
\suppressfloats[t]

\subsection{Training \method}
\label{app:train:ours}

\textbf{Policy optimization.} The policy is Qwen2.5-VL-3B-Instruct, trained full-parameter with
OpenRLHF on four GPUs (bf16, ZeRO-2, gradient checkpointing, FlashAttention-2). The advantage
estimator is RLOO with $G=8$ captions per chart at temperature $1.0$, $32$ charts per rollout,
training batch size $128$, learning rate $10^{-6}$ with $20$ warm-up steps, KL coefficient $0$,
reward normalization, generation capped at $1{,}024$ tokens and no format reward. Four passes over
the $9{,}500$ training charts amount to about $1{,}200$ rollout steps. Every $50$ steps we caption the
$500$ validation charts greedily and score them against their fixed distractors; validation reward
peaks at step $550$ ($1.808$, verifier accuracy $90.4\%$, against $1.172$ and $58.6\%$ at
initialization) and is flat afterwards, so step $550$ is the reported checkpoint.

\textbf{Reward.} The frozen verifier is Qwen2.5-14B-Instruct in non-thinking mode, served alongside
training under the prompt of Appendix~\ref{app:prompt:verifier}. The candidate set has $K=4$ members printed as raw numbers at a shared precision, the
reward scale is $\lambda=2$, so a caption earns one of $\{0,0.5,1,1.5,2\}$, and the overlap filter of \S\ref{sec:method:ours} uses $\eta=0.2$. The
same caption instruction (Appendix~\ref{app:prompt:caption}) is used at rollout, validation and evaluation time.

\subsection{Baselines}
\label{app:baselines}

All four fine-tuned captioners start from the same Qwen2.5-VL-3B-Instruct weights, are trained
full-parameter, read the same charts and receive the same caption instruction
(Appendix~\ref{app:prompt:caption}). The three RL models share the policy optimizer of
Appendix~\ref{app:train:ours}: RLOO, $G=8$ captions per chart at temperature $1.0$, $32$ charts per
rollout, training batch $128$, learning rate $10^{-6}$ with $20$ warm-up steps, KL coefficient
$0$, generation capped at $1{,}024$ tokens, no format reward, bf16 with ZeRO-2 on four GPUs.
Table~\ref{tab:arms} lists what differs, and the paragraphs below give the reward of each
baseline.

\begin{table}[t]
\centering
\caption{Configuration of the four fine-tuned models. Each model is selected on the
$500$ validation charts under its own training signal, so no model is selected on a test
metric. The answerability model trains on fewer charts because a chart enters its training
set only if at least two of its questions survive verification.}
\label{tab:arms}
\small
\setlength{\tabcolsep}{4pt}
\begin{tabular}{@{}>{\raggedright\arraybackslash}p{2.2cm}>{\raggedright\arraybackslash}p{2.45cm}>{\raggedright\arraybackslash}p{2.65cm}>{\raggedright\arraybackslash}p{2.45cm}>{\raggedright\arraybackslash}p{2.45cm}@{}}
\toprule
& \method & RL, answerability & RL, LLM-as-judge & SFT \\
\midrule
Training signal & identification among $K=4$ candidates & accuracy on the chart's verified questions & grade $1$--$10$ of the caption & cross-entropy on 72B captions \\
Frozen model in the loop & Qwen2.5-14B-Instruct verifier & Qwen2.5-14B-Instruct answerer & Qwen2.5-14B-Instruct judge & Qwen2.5-VL-72B-Instruct teacher \\
Training charts & $9{,}500$ & $7{,}243$ ($23{,}436$ questions) & $9{,}500$ & $9{,}500$ pairs \\
Passes over the data & $4$ & $8$ & $2$ & $5$ \\
Learning rate & $10^{-6}$ & $10^{-6}$ & $10^{-6}$ & $10^{-5}$, cosine \\
Checkpoint every & $50$ steps & $25$ steps & $50$ steps & $50$ steps \\
Selection rule & highest validation reward & highest validation reward & highest validation reward & lowest validation loss \\
\bottomrule
\end{tabular}
\end{table}

\textbf{RL, answerability.} This baseline ports the CapRL recipe \citep{xing2025caprl}: a VLM writes
multiple-choice questions about each training chart, and the reward is the accuracy of a frozen
text-only answerer that sees the caption alone. Questions are written by qwen-vl-max reading the
chart alone, at temperature $0.8$, ten four-option questions with keys per chart under the
instruction of Appendix~\ref{app:prompt:qagen}, which is CapRL's question prompt adapted to a
series: it asks for questions about the shape and where its features sit and forbids exact
values, since a chart without value labels cannot support them. This yields $95{,}329$
questions over $9{,}985$ charts. Because a VLM confidently answers its own questions wrongly,
every question is then checked against the raw values: a text LLM (Qwen3.6-27B) drives a
tool-calling loop that measures the series (extrema, segment statistics, feature detection) and
the check itself is done by code, and the LLM only drives the tools. Only $27.9\%$ of the questions
($26{,}570$ over $9{,}332$ charts) survive with the key the VLM gave; a further $21.5\%$ would
need their key corrected and are discarded together with the $50.6\%$ that are ambiguous, not
computable from the values, or wrong. Of the surviving questions, $99.4\%$ have four time
intervals as options. Charts with fewer than two surviving questions are dropped, leaving
$7{,}243$ training charts with $23{,}436$ questions and $500$ validation charts with $1{,}545$
questions. The answerer is the same frozen Qwen2.5-14B-Instruct that serves as our verifier; it is
queried once per question under the
prompt of Appendix~\ref{app:prompt:answerer} at temperature $0.6$, and the reward is the fraction
of the chart's questions it answers correctly, scaled to $[0,2]$ like ours. Training follows the
shared recipe with eight passes and a checkpoint every $25$ steps; the checkpoint with the highest
validation reward, step $650$ ($1.238$, accuracy $61.9\%$, against $0.574$ and $28.7\%$ at
initialization), is the reported one.

\textbf{RL, LLM-as-judge.} The judge is again the frozen Qwen2.5-14B-Instruct that serves as
our verifier, so that the three RL models differ in the form of the reward and not in the model
behind it. It receives the true series as raw values and the caption, never the chart or any
other series, and grades the caption from $1$ to $10$ under the rubric of
Appendix~\ref{app:prompt:judge}, which weighs accuracy above specificity above completeness, pins
an empty or generic caption to $1$, and instructs the judge not to reward length; the reward is
$(g-1)/9\times2$, and an unparseable reply counts as $1$ so that the RLOO group stays complete.
No length or format penalty is added beyond the rubric. Under the shared recipe with two passes
(about $600$ steps) the reward saturates: from step $410$ the mean group reward stays at $2.0$, and
over the last $100$ steps the within-group standard deviation is $0.004$, so the policy gradient
vanishes. Validation reward reaches its ceiling at steps $500$ and $550$, and step $550$ is the
reported checkpoint. Its captions show the hack that Figure~\ref{fig:case} illustrates:
$99.9\%$ of them end with a paragraph addressed to the grader, which takes $43.8\%$ of their
characters; the corresponding share is $12.1\%$ for the untuned initialization and $0.0\%$ for
\method.

\textbf{SFT on the 72B teacher's captions.} The teacher Qwen2.5-VL-72B-Instruct captions the
same $9{,}500$ training and $500$ validation charts under the same caption instruction with
greedy decoding; its captions have a median length of $2{,}283$ characters and $85.1\%$ of them
open with the same eight words. Captions shorter than $80$ characters or without sentence-final
punctuation are dropped, which removes none of the $9{,}500$ training pairs and one validation
pair. The student is trained full-parameter with the \texttt{trl} SFT trainer for five epochs
($742$ steps) at learning rate $10^{-5}$ with a cosine schedule and $5\%$ warm-up, effective
batch $64$, bf16, sequence length $2{,}048$, evaluating on the $499$ validation pairs every $50$
steps; validation loss decreases monotonically to $0.199$ at step $700$, which is the reported
checkpoint. Scored under our reward on the validation charts, the teacher's own captions reach an
identification accuracy of $68.3\%$, between the untuned 3B student ($58.6\%$) and \method\
($90.4\%$), so imitating them cannot go beyond that level.

\subsection{Ablation settings}
\label{app:train:ablation}

Each ablation setting changes one component of the recipe of Appendix~\ref{app:train:ours} and
keeps everything else: the training data, the caption instruction, the policy-optimization
hyperparameters and the Criterion 3 predictor. The reward-model ladder replaces the Qwen2.5-14B verifier
of the reward with Qwen2.5-7B or Qwen2.5-3B, so on this ladder the model that scores the reward
changes while the evaluation verifiers of Appendix~\ref{app:eval} do not. The candidate ladder
raises $K$ from $4$ to $6$ or $8$ with the same 14B verifier; the $K-1$ distractors of a smaller
$K$ are a prefix of those of a larger one, so raising $K$ adds candidates rather than exchanging
them. The distractor-rule setting keeps $K=4$ and the 14B verifier and replaces the
nearest-neighbor rule by a uniform random draw from series of the same length. Every setting is a
single training run at one seed (Appendix~\ref{app:sig}) on the learning-rate schedule of
\method, so the settings of a ladder differ only in the component named above.

\section{Evaluation details}
\label{app:eval}

\textbf{Overlap between training and evaluation data.}
\label{app:overlap}
The $10{,}000$ training fragments carry no source label, so we matched their values against the
evaluation series directly. $6{,}147$ fragments are windows of the OT column of ETTh1 or ETTm1, the
two ETT sites that TSFragment-600K draws on; none is a window of any column of ETTh2 or ETTm2
(every fragment was checked against every column of both files at a tolerance of $0.011$, the
$57$ near-constant fragments at every offset), and neither Saugeen nor the Australian demand
corpus is a source of TSFragment-600K. In the other direction, no BEDTime or CaTS-Bench series
shares a run of $12$ or more consecutive values with a training fragment, apart from flat
stretches of at most two distinct values ($5$ series on BEDTime, $72$ on CaTS-Bench) and one
$14$-step air-quality stretch whose values lie between $0.03$ and $0.05$, inside the matching
tolerance. ETTh2 and ETTm2 thus share a domain and a variable with part of the training data,
but no observation.

\subsection{Criterion 1: entailment}

BEDTime supplies $2{,}000$ series whose references are crowd-sourced or
template-generated rather than LLM-written, giving $|\mathcal{P}|=3{,}958$ pairs; we use DeBERTa
long-NLI at $\tau=0.5$; an empty caption scores $0.003$ on this criterion and a reference scored
against itself $1.000$. CaTS-Bench ships one reference per series but none of it is usable as written, since it
states calendar dates, the place, the phenomenon and statistics of the source dataset, none of
which our charts carry. The reported CaTS-Bench column therefore scores against rewritten
references, described next.

\textbf{Rewriting the CaTS-Bench references.}
\label{app:cats-rewrite}
The $2{,}969$ references were rewritten in two passes by Claude Opus 5, which saw the reference
text only, never the series, the chart or any metadata; each pass was split into $30$ slices of
about $100$ references, every slice under the identical instruction, and a rewrite that broke a
rule on the first attempt was redone under the same instruction. Pass~1 removes the domain: the
phenomenon or attribute name, places, units and currencies, comparisons to statistics outside the
series, and calendar dates, which become relative positions (``about two-thirds of the way
through''); it keeps every observation in its order and every number as written
(Appendix~\ref{app:prompt:cats1}). Pass~2 starts from the output of pass~1 and removes every
numeral and quantity, replacing each by the comparative fact it supported (``starting at 91.29
and reaching 105.27'' becomes ``starting low and ending higher''), so that the two reference
sets differ only in the numerals (Appendix~\ref{app:prompt:cats2}). A script checks the result:
after pass~1 every reference still contains a digit, after pass~2 none of the $2{,}969$ does, and
the mean length falls from $255$ to $220$ characters. We report the pass-2 references because
$99.9\%$ of the pass-1 references state an exact quantity against $1.3\%$ of BEDTime's, and a
caption written from a chart without value labels cannot entail an exact quantity however well
it describes the shape: scored against pass~1, every chart captioner lands at the empty-caption
floor, and the column would rank models by their access to the digits instead of by caption quality.

\subsection{Criterion 2: identification}

Evaluation series come from BEDTime ($1{,}500$ series from three of its four
public subsets; \textsc{sushi} is excluded because its $2{,}048$-point series need roughly $57$k
tokens for a single four-way option set, past the verifier's context) and CaTS-Bench ($2{,}969$
series, lengths $8$--$150$). We use their series but build our own protocol, because both benchmarks test what a model can
do with the series itself, and we test what the caption preserves. We
build the candidate sets ourselves, at $K=4$, as the three nearest neighbors in standardized
$(\mathrm{mean},\mathrm{std},\min,\max)$ space at the same length, whereas BEDTime's own distractors
are selected to be maximally dissimilar, which its authors describe as an upper bound on
performance. Each identification number is the mean over two frozen verifiers that took no part
in training any model, GLM-4-9B-0414 and Phi-4 (14B), each queried at all four gold positions
(next paragraph). Per-position accuracy is reported as a control, and all captioners answer the
same items.

\textbf{On the choice of the evaluation verifiers.}
\label{app:verifier-sharing}
The reward verifier is Qwen2.5-14B-Instruct, and \textsc{cap}$\to$\textsc{ser} asks the
question the reward asks. If the same model scored it, that column would only show how far the training
objective generalizes to new series and candidate sets. The identification columns of Table~\ref{tab:criteria12}
are therefore scored by two frozen LLMs that took no part in training any model, GLM-4-9B-0414
and Phi-4 (14B), and report their mean. Tables~\ref{tab:xv-bedtime} and~\ref{tab:xv-cats} give
every captioner's accuracy under five verifiers, the reward verifier, the two evaluation
verifiers and two further LLMs, all under the same four-rotation protocol, so the reading of
any one verifier can be checked against the others; Appendix~\ref{app:results:verifiers} says
why the two further LLMs are not averaged into the main table. Three checks show that the verifier is not driving the result. No verifier answers without
the caption: with the caption
replaced by a placeholder, the two evaluation verifiers identify the true series among the
nearest-neighbor candidates at $0.18$ to $0.29$ (chance $0.25$), whereas in the pilot that chose
the distractor rule (\S\ref{sec:method:ours}) the shuffled and reversed distractors we rejected
are identified at $72.5\%$ without a caption. \textsc{ser}$\to$\textsc{cap} is a
direction the reward never asks, Criterion 1 is scored by an NLI model against reference
captions and Criterion 3 by a caption-only predictor, neither of which is the verifier, and
\method\ leads on all of them.
On reference captions that none of the captioners wrote, the 14B reward verifier is more accurate than
the 7B and 3B verifiers on all three domains and at every $K$ (Figure~\ref{fig:ablation}c,
Appendix~\ref{app:ablation}), so a larger verifier is simply more accurate, not more partial to
one captioner's style.

\subsection{Criterion 3: caption-only forecasting and reconstruction}

The two targets share the decoder class and the windows, and in both cases
the training loss and the reported metric are the same quantity, so no sampling noise enters the
measurement. The split is chronological, so that adjacent windows cut at stride $1$
cannot land on both sides of it. Each corpus is one univariate series whose rows are divided
$60/20/20$ in time order into a training, a validation and a test segment, following the ETT
convention; a window is assigned to the segment that contains the first step of its forecasting
target, which is the step after its lookback, and a window whose target runs past the end of the
series is discarded. Every test target therefore starts later than every validation target, and
every validation target later than every training target. The only contact between segments is
at their boundary, exactly as in the standard forecasting protocol: the lookback of the earliest
windows of a segment reaches back into the segment before it, and a training target runs at most
$15$ steps past the training segment. A test window never shares a step with any training
window or training target, since the validation segment is longer than a window on every corpus.
Reconstruction uses the same windows and the same assignment, so its test targets are the
lookback windows of the test split, which reach into the validation segment but never into the
training segment. The predictor is fitted on the training windows, early-stopped on the validation
windows, and reported on the test windows only. Values are standardized with the statistics of the
training segment, which makes $\widehat{\mathcal{S}}$ dimensionless and comparable across datasets. The
text-blind predictor $b$, the naive baseline of Table~\ref{tab:readout}, is last-value persistence for
forecasting and the mean training window for reconstruction; the empty-caption and detachment
controls of \S\ref{sec:method:eval} are the empty and shuffled caption rows of that table. On truncation:
caption length differs across captioners by more than $2\times$, so the encoder's default $256$-token cap
would have cut every caption of the more verbose captioners and $21\%$ of ours, and the
ranking would then partly reflect length. Every run records its token statistics and aborts if any
caption would be cut.

\textbf{Caption-only predictor.} Windows are cut with a lookback of $128$ and, for forecasting, a horizon of $16$,
at stride $1$, assigned to the chronological $60/20/20$ segments described above,
and subsampled uniformly to $5{,}000$ windows per dataset under a fixed data seed, which leaves
about $3{,}000$ training, $1{,}000$ validation and $1{,}000$ test windows per corpus, the same
windows for every captioner. Values are standardized with the statistics of the training
segment. The decoder is \texttt{google/embeddinggemma-300m} with a LoRA
adapter ($r=16$), mean-pooled and fed to an MLP head, trained for up to $30$ epochs at learning rate
$3\times10^{-4}$, keeping the epoch with the lowest validation MSE and stopping after $6$ epochs
without improvement; the test windows are scored once, with that epoch's weights. The recipe is held fixed across captioners, so differences between cells come from the captions
and not from tuning, and every cell is run with five training seeds.
Table~\ref{tab:readout} reports the mean over those seeds; Table~\ref{tab:readout-std}
repeats every cell with its standard deviation.

\subsection{Aggregation for the ablation figure and Table~\ref{tab:negrand}}

\textbf{Aggregation.} Figure~\ref{fig:ablation}(a,b) and Table~\ref{tab:negrand} aggregate
the per-setting tables of Appendix~\ref{app:results:ablation} as follows. Every caption-level number is the unweighted mean of its BEDTime and
CaTS-Bench values, except the identification bar of the figure, which is
\textsc{ser}$\to$\textsc{cap} on CaTS-Bench alone; the BEDTime identification of every setting is in
Tables~\ref{tab:abl_rm_caption} and~\ref{tab:abl_neg_caption}, and Table~\ref{tab:negrand} lists
both directions averaged over the two benchmarks. The forecasting and reconstruction numbers are
geometric means over the four corpora of the ratio of a setting's test MSE to \method's, so that
corpora on different error scales weigh equally; the figure plots the inverse so that taller is
better. The reward row of Table~\ref{tab:negrand} is the identification accuracy above chance,
$(\mathrm{acc}-1/K)/(1-1/K)$, reached on the setting's own candidate sets at the reported checkpoint.

\subsection{Statistical tests and noise}
\label{app:sig}

\textbf{Criteria 1 and 2.} Both criteria are deterministic functions of a caption set: captions
are decoded greedily, the candidate sets and the NLI threshold are fixed, and every captioner
answers the same items. Tables~\ref{tab:criteria12}, \ref{tab:xv-bedtime} and~\ref{tab:xv-cats}
therefore report each number once, as is standard for benchmark tables, and bold marks the best
value of the fine-tuned group. The two statements about SFT in \S\ref{sec:intro} are read off
Table~\ref{tab:criteria12}: on BEDTime entailment SFT scores $0.281$ against $0.296$ for the
untuned initialization, so SFT does not raise entailment there, and we do not claim that it
lowers it; on CaTS-Bench \textsc{ser}$\to$\textsc{cap} SFT is below the initialization under
both evaluation verifiers ($0.598$ against $0.680$ under GLM-4-9B, $0.634$ against $0.763$ under
Phi-4-14B).

\textbf{Significance of the Criterion 3 gaps.} Bolding in Table~\ref{tab:readout} comes from a two-sided paired $t$-test across the
5 predictor seeds. Seeds are paired because a seed moves every captioner together: only the
predictor's training loop varies, so all 5 runs score identical windows against identical
captions, and the paired difference removes the shared per-seed component. In each column we take
the captioner with the lowest mean and bold it together with every captioner it cannot be separated from at
$p<0.05$, so two bold entries mark a tie rather than a ranking. Table~\ref{tab:readout-sig} lists
the $p$-values of \method\ against every other captioner. With $n=5$
($\mathrm{df}=4$) the test is lenient; it only tells which gaps are stable across seeds.

\begin{table}[t]
\centering
\caption{Paired $t$-test $p$-values, \method\ against every other captioner.
Two-sided, paired across the 5 predictor seeds. \method\ has the lower mean everywhere
except the cells marked $\dagger$, where the other captioner is ahead.}
\label{tab:readout-sig}
\small
\setlength{\tabcolsep}{6pt}
\begin{tabular}{@{}lccccc@{}}
\toprule
& SFT & Answ. & Judge & 72B & 3B init \\
\midrule
\multicolumn{6}{@{}l}{\emph{Forecasting}} \\
~~ETTh2 & 0.001 & $<$0.001 & 0.002 & 0.002 & $<$0.001 \\
~~ETTm2 & 0.016 & $<$0.001 & 0.035 & $<$0.001$^{\dagger}$ & $<$0.001 \\
~~Saug. & 0.007 & $<$0.001 & 0.030 & 0.005 & $<$0.001 \\
~~Elec. & $<$0.001 & $<$0.001 & $<$0.001 & $<$0.001 & $<$0.001 \\
\multicolumn{6}{@{}l}{\emph{Reconstruction}} \\
~~ETTh2 & $<$0.001 & $<$0.001 & $<$0.001 & 0.008 & $<$0.001 \\
~~ETTm2 & 0.010 & $<$0.001 & 0.007 & 0.609$^{\dagger}$ & $<$0.001 \\
~~Saug. & $<$0.001 & $<$0.001 & 0.088 & 0.678$^{\dagger}$ & $<$0.001 \\
~~Elec. & $<$0.001 & $<$0.001 & $<$0.001 & $<$0.001 & $<$0.001 \\
\bottomrule
\end{tabular}
\end{table}

Two further families of tests back the claims in \S\ref{sec:method:eval}. Every fine-tuned
captioner beats both text-only controls in all eight dataset-study cells: across those
64 comparisons the weakest is $p=9.8\times 10^{-5}$. The two controls are in turn
indistinguishable from each other in 7 of the $8$ cells, the exception being forecasting on
ETTm2 ($p=0.003$), where the
shuffled caption is $0.5\%$ worse than no caption at all. A wrong caption is therefore about as useful as
none, so the margin of a captioner over the controls measures series-specific content.

\begin{table}[t]
\centering
\caption{Criterion 3 with across-seed spread. The cells of
Table~\ref{tab:readout}, laid out one study per block so the two panels can be read
separately. Bolding and baselines are as in that table. Only the predictor's training
loop changes between seeds, so the spread comes from the predictor; each captioner is a single
RL or SFT run.}
\label{tab:readout-std}
\footnotesize
\setlength{\tabcolsep}{4pt}
\begin{tabular}{@{}lcccc@{}}
\toprule
& \multicolumn{4}{c}{\textbf{Forecasting} (predict the next window)} \\
\cmidrule(lr){2-5}
Captioner & ETTh2 & ETTm2 & Saug. & Elec. \\
\midrule
Naive  baseline & 0.3047 & 0.0596 & 1.2955 & 0.9078 \\
Empty caption & 0.8446 {\tiny$\pm$0.0101} & 0.8001 {\tiny$\pm$0.0042} & 1.0601 {\tiny$\pm$0.0016} & 0.9129 {\tiny$\pm$0.0041} \\
Shuffled caption & 0.8437 {\tiny$\pm$0.0087} & 0.8040 {\tiny$\pm$0.0046} & 1.0619 {\tiny$\pm$0.0042} & 0.9124 {\tiny$\pm$0.0043} \\
\addlinespace[2pt]
Qwen2.5-VL-3B & 0.3062 {\tiny$\pm$0.0086} & 0.1012 {\tiny$\pm$0.0047} & 0.9419 {\tiny$\pm$0.0119} & 0.5751 {\tiny$\pm$0.0351} \\
Qwen2.5-VL-72B & 0.2149 {\tiny$\pm$0.0094} & 0.0451 {\tiny$\pm$0.0009} & 0.8521 {\tiny$\pm$0.0061} & 0.3089 {\tiny$\pm$0.0094} \\
\midrule
\multicolumn{5}{@{}l}{\emph{Fine-tuned from Qwen2.5-VL-3B}} \\
SFT on 72B's captions & 0.2595 {\tiny$\pm$0.0151} & 0.0615 {\tiny$\pm$0.0021} & 0.9026 {\tiny$\pm$0.0216} & 0.3480 {\tiny$\pm$0.0086} \\
RL, answerability & 0.3780 {\tiny$\pm$0.0168} & 0.2110 {\tiny$\pm$0.0159} & 0.9384 {\tiny$\pm$0.0104} & 0.3258 {\tiny$\pm$0.0151} \\
RL, LLM-as-judge & 0.2384 {\tiny$\pm$0.0134} & 0.0630 {\tiny$\pm$0.0077} & 0.8619 {\tiny$\pm$0.0217} & 0.3388 {\tiny$\pm$0.0113} \\
\method & \textbf{0.1806} {\tiny$\pm$0.0078} & \textbf{0.0541} {\tiny$\pm$0.0022} & \textbf{0.8175} {\tiny$\pm$0.0196} & \textbf{0.2042} {\tiny$\pm$0.0091} \\
\bottomrule
\end{tabular}

\vspace{9pt}

\begin{tabular}{@{}lcccc@{}}
\toprule
& \multicolumn{4}{c}{\textbf{Reconstruction} (predict the window itself)} \\
\cmidrule(lr){2-5}
Captioner & ETTh2 & ETTm2 & Saug. & Elec. \\
\midrule
Naive  baseline & 0.9501 & 0.8335 & 1.1748 & 0.8634 \\
Empty caption & 0.9284 {\tiny$\pm$0.0306} & 0.8274 {\tiny$\pm$0.0029} & 1.1764 {\tiny$\pm$0.0008} & 0.8667 {\tiny$\pm$0.0021} \\
Shuffled caption & 0.9347 {\tiny$\pm$0.0121} & 0.8270 {\tiny$\pm$0.0015} & 1.1774 {\tiny$\pm$0.0018} & 0.8663 {\tiny$\pm$0.0010} \\
\addlinespace[2pt]
Qwen2.5-VL-3B & 0.2445 {\tiny$\pm$0.0052} & 0.0633 {\tiny$\pm$0.0029} & 0.6852 {\tiny$\pm$0.0173} & 0.4352 {\tiny$\pm$0.0156} \\
Qwen2.5-VL-72B & 0.1134 {\tiny$\pm$0.0065} & 0.0288 {\tiny$\pm$0.0022} & 0.4444 {\tiny$\pm$0.0113} & 0.1993 {\tiny$\pm$0.0077} \\
\midrule
\multicolumn{5}{@{}l}{\emph{Fine-tuned from Qwen2.5-VL-3B}} \\
SFT on 72B's captions & 0.1478 {\tiny$\pm$0.0044} & 0.0333 {\tiny$\pm$0.0016} & 0.5259 {\tiny$\pm$0.0177} & 0.2144 {\tiny$\pm$0.0084} \\
RL, answerability & 0.3391 {\tiny$\pm$0.0185} & 0.1883 {\tiny$\pm$0.0151} & 0.5309 {\tiny$\pm$0.0091} & 0.2608 {\tiny$\pm$0.0095} \\
RL, LLM-as-judge & 0.1373 {\tiny$\pm$0.0104} & 0.0325 {\tiny$\pm$0.0019} & \textbf{0.4517} {\tiny$\pm$0.0027} & 0.2248 {\tiny$\pm$0.0029} \\
\method & \textbf{0.0989} {\tiny$\pm$0.0033} & \textbf{0.0294} {\tiny$\pm$0.0010} & \textbf{0.4463} {\tiny$\pm$0.0054} & \textbf{0.1357} {\tiny$\pm$0.0068} \\
\bottomrule
\end{tabular}

\vspace{3pt}
{\footnotesize\raggedright Seeds 2020, 2021, 2022, 2023, 2024; all 5 score identical windows
against identical captions.\par}
\end{table}

\textbf{Number of training runs.}
\label{app:runs}
Every fine-tuned captioner in Tables~\ref{tab:criteria12} and~\ref{tab:readout} and every
ablation setting of \S\ref{sec:exp:ablation} is a single training run at one seed; only the
Criterion 3 predictor is repeated over seeds, so the spreads in Table~\ref{tab:readout} are the
predictor's, not the captioner's. Two noise scales help read the gaps.
Scoring one checkpoint twice moves a Criterion 2 accuracy by about $\pm0.005$, and across the
five predictor seeds the relative standard deviation of a Table~\ref{tab:readout} cell has median
$3.6\%$ and $90$th percentile $6.1\%$. \method's margin over the best fine-tuned baseline is
between $+0.114$ and $+0.253$ on the six columns of Table~\ref{tab:criteria12} and above $9\%$ of
the error in six of the eight cells of Table~\ref{tab:readout}; the two Saugeen cells are $5\%$
(forecasting) and $1\%$ (the reconstruction tie). The
Criterion 3 tests compare the caption sets we obtained; they do not cover the variance of
training itself, and a rerun of any model could move its numbers by more than the evaluation
noise. The test script is part of the released code.

\section{Additional results}
\label{app:ablation}

\subsection{Identification under five verifiers}
\label{app:results:verifiers}
Tables~\ref{tab:xv-bedtime} and~\ref{tab:xv-cats} re-score every captioner of
Table~\ref{tab:criteria12} under five frozen verifiers: the reward verifier, the two evaluation
verifiers whose mean the main table reports, and two further LLMs. Everything except the
verifier is as in Appendix~\ref{app:eval}.

The main table averages GLM-4-9B and Phi-4-14B and not the other two because a verifier is only
informative where it separates captions. The reward verifier is set aside for the reason given in
Appendix~\ref{app:verifier-sharing}. Llama-3.1-8B and Mistral-Nemo-12B score every VLM captioner
below $0.45$ on \textsc{s}$\to$\textsc{c} on both benchmarks, and the three fine-tuned baselines
between $0.26$ and $0.33$ on BEDTime, close to chance ($0.25$), so in that direction their
readings say more about the verifier than about the captions; GLM-4-9B and Phi-4-14B place the
same three baselines between $0.33$ and $0.50$ on BEDTime and between $0.52$ and $0.65$ on CaTS-Bench.
The two are still worth reporting as a check on the direction they do resolve: on
\textsc{c}$\to$\textsc{s}, \method\ leads the fine-tuned models under all five verifiers on
both benchmarks.

\begin{table}[t]
\centering
\caption{Identification on BEDTime under five frozen verifiers. Every captioner's
\textsc{c}$\to$\textsc{s} and \textsc{s}$\to$\textsc{c} accuracy on the candidate sets of
Table~\ref{tab:criteria12}, re-scored by each verifier under the same four-rotation protocol
(chance $0.25$). Qwen2.5-14B is the reward verifier; GLM-4-9B and Phi-4-14B are the two evaluation
verifiers whose mean Table~\ref{tab:criteria12} reports. Bold marks the best of the four models
fine-tuned from the same initialization, as in Table~\ref{tab:criteria12}.}
\label{tab:xv-bedtime}
\footnotesize
\setlength{\tabcolsep}{2.6pt}
\resizebox{\textwidth}{!}{%
\begin{tabular}{@{}lcccccccccc@{}}
\toprule
& \multicolumn{2}{c}{Qwen2.5-14B} & \multicolumn{2}{c}{GLM-4-9B} & \multicolumn{2}{c}{Llama-3.1-8B} & \multicolumn{2}{c}{Mistral-Nemo-12B} & \multicolumn{2}{c}{Phi-4-14B} \\
\cmidrule(lr){2-3} \cmidrule(lr){4-5} \cmidrule(lr){6-7} \cmidrule(lr){8-9} \cmidrule(lr){10-11}
Captioner & \textsc{c}$\to$\textsc{s} & \textsc{s}$\to$\textsc{c} & \textsc{c}$\to$\textsc{s} & \textsc{s}$\to$\textsc{c} & \textsc{c}$\to$\textsc{s} & \textsc{s}$\to$\textsc{c} & \textsc{c}$\to$\textsc{s} & \textsc{s}$\to$\textsc{c} & \textsc{c}$\to$\textsc{s} & \textsc{s}$\to$\textsc{c} \\
\midrule
Qwen2.5-VL-3B & 0.664 & 0.566 & 0.559 & 0.443 & 0.472 & 0.328 & 0.518 & 0.349 & 0.541 & 0.548 \\
Qwen2.5-VL-72B & 0.742 & 0.625 & 0.610 & 0.455 & 0.582 & 0.338 & 0.574 & 0.334 & 0.600 & 0.513 \\
Qwen3-VL-8B & 0.752 & 0.604 & 0.639 & 0.446 & 0.648 & 0.372 & 0.620 & 0.351 & 0.700 & 0.549 \\
InternVL3-14B & 0.709 & 0.600 & 0.585 & 0.467 & 0.542 & 0.360 & 0.541 & 0.324 & 0.539 & 0.557 \\
Phi-3.5-mini & 0.936 & 0.903 & 0.893 & 0.828 & 0.884 & 0.668 & 0.837 & 0.701 & 0.925 & 0.787 \\
Qwen2.5-7B & 0.948 & 0.926 & 0.904 & 0.835 & 0.906 & 0.599 & 0.849 & 0.604 & 0.928 & 0.812 \\
Qwen2.5-14B & 0.947 & 0.937 & 0.901 & 0.850 & 0.909 & 0.725 & 0.870 & 0.645 & 0.937 & 0.825 \\
ChatTS-14B & 0.757 & 0.775 & 0.744 & 0.659 & 0.728 & 0.487 & 0.637 & 0.564 & 0.622 & 0.782 \\
\midrule
\multicolumn{11}{@{}l}{\emph{Fine-tuned from Qwen2.5-VL-3B}} \\
SFT on 72B's captions & 0.714 & 0.588 & 0.582 & 0.420 & 0.534 & 0.325 & 0.547 & \textbf{0.324} & 0.559 & 0.476 \\
RL, answerability & 0.624 & 0.499 & 0.453 & 0.390 & 0.384 & 0.279 & 0.455 & 0.281 & 0.443 & 0.495 \\
RL, LLM-as-judge & 0.718 & 0.549 & 0.534 & 0.331 & 0.341 & 0.255 & 0.518 & 0.264 & 0.554 & 0.493 \\
\method & \textbf{0.788} & \textbf{0.765} & \textbf{0.717} & \textbf{0.596} & \textbf{0.613} & \textbf{0.340} & \textbf{0.612} & \textbf{0.324} & \textbf{0.700} & \textbf{0.680} \\
\bottomrule
\end{tabular}}
\end{table}

\begin{table}[t]
\centering
\caption{Identification on CaTS-Bench under five frozen verifiers. Every captioner's
\textsc{c}$\to$\textsc{s} and \textsc{s}$\to$\textsc{c} accuracy on the candidate sets of
Table~\ref{tab:criteria12}, re-scored by each verifier under the same four-rotation protocol
(chance $0.25$). Qwen2.5-14B is the reward verifier; GLM-4-9B and Phi-4-14B are the two evaluation
verifiers whose mean Table~\ref{tab:criteria12} reports. Bold marks the best of the four models
fine-tuned from the same initialization, as in Table~\ref{tab:criteria12}.}
\label{tab:xv-cats}
\footnotesize
\setlength{\tabcolsep}{2.6pt}
\resizebox{\textwidth}{!}{%
\begin{tabular}{@{}lcccccccccc@{}}
\toprule
& \multicolumn{2}{c}{Qwen2.5-14B} & \multicolumn{2}{c}{GLM-4-9B} & \multicolumn{2}{c}{Llama-3.1-8B} & \multicolumn{2}{c}{Mistral-Nemo-12B} & \multicolumn{2}{c}{Phi-4-14B} \\
\cmidrule(lr){2-3} \cmidrule(lr){4-5} \cmidrule(lr){6-7} \cmidrule(lr){8-9} \cmidrule(lr){10-11}
Captioner & \textsc{c}$\to$\textsc{s} & \textsc{s}$\to$\textsc{c} & \textsc{c}$\to$\textsc{s} & \textsc{s}$\to$\textsc{c} & \textsc{c}$\to$\textsc{s} & \textsc{s}$\to$\textsc{c} & \textsc{c}$\to$\textsc{s} & \textsc{s}$\to$\textsc{c} & \textsc{c}$\to$\textsc{s} & \textsc{s}$\to$\textsc{c} \\
\midrule
Qwen2.5-VL-3B & 0.873 & 0.803 & 0.674 & 0.680 & 0.675 & 0.429 & 0.591 & 0.449 & 0.827 & 0.763 \\
Qwen2.5-VL-72B & 0.951 & 0.806 & 0.711 & 0.631 & 0.763 & 0.372 & 0.648 & 0.434 & 0.896 & 0.655 \\
Qwen3-VL-8B & 0.933 & 0.775 & 0.716 & 0.587 & 0.802 & 0.379 & 0.650 & 0.413 & 0.919 & 0.665 \\
InternVL3-14B & 0.919 & 0.798 & 0.684 & 0.689 & 0.734 & 0.384 & 0.619 & 0.426 & 0.849 & 0.713 \\
Phi-3.5-mini & 0.989 & 0.988 & 0.979 & 0.953 & 0.978 & 0.819 & 0.867 & 0.849 & 0.981 & 0.826 \\
Qwen2.5-7B & 0.997 & 0.991 & 0.982 & 0.961 & 0.981 & 0.753 & 0.869 & 0.850 & 0.988 & 0.839 \\
Qwen2.5-14B & 0.999 & 0.997 & 0.987 & 0.970 & 0.987 & 0.881 & 0.915 & 0.859 & 0.996 & 0.906 \\
ChatTS-14B & 0.889 & 0.923 & 0.871 & 0.823 & 0.864 & 0.667 & 0.717 & 0.712 & 0.893 & 0.904 \\
\midrule
\multicolumn{11}{@{}l}{\emph{Fine-tuned from Qwen2.5-VL-3B}} \\
SFT on 72B's captions & 0.919 & 0.759 & 0.673 & 0.598 & 0.722 & 0.362 & 0.605 & \textbf{0.414} & 0.846 & 0.634 \\
RL, answerability & 0.746 & 0.627 & 0.523 & 0.523 & 0.507 & 0.291 & 0.478 & 0.311 & 0.675 & 0.597 \\
RL, LLM-as-judge & 0.915 & 0.725 & 0.650 & 0.529 & 0.535 & 0.275 & 0.561 & 0.319 & 0.738 & 0.650 \\
\method & \textbf{0.964} & \textbf{0.951} & \textbf{0.868} & \textbf{0.846} & \textbf{0.810} & \textbf{0.431} & \textbf{0.713} & 0.400 & \textbf{0.904} & \textbf{0.891} \\
\bottomrule
\end{tabular}}
\end{table}

\subsection{Reward discrimination on reference captions}
\label{app:results:grid}

Figure~\ref{fig:ablation}(c) asks each of
the three verifiers the \textsc{cap}$\to$\textsc{ser} question at $K\in\{4,6,8,10,12\}$ on
captions that none of the captioners wrote: the $500$ validation charts of the training distribution with the
GPT-4o-mini captions shipped with TSFragment-600K, $2{,}613$ CaTS-Bench series with the
number-free references of Appendix~\ref{app:cats-rewrite}, and $1{,}480$ BEDTime series with their
first annotation. Candidates are the nearest neighbors in the same statistic space as the reward,
the true series is rotated through every position, and the items are shared across verifiers and
$K$, so the only variable is the verifier and the number of candidates. Absolute values are lower
than on the policy's captions, which the reward has optimized to be discriminable; what matters is
the ordering, which is $14$B $>$ $7$B $>$ $3$B on every domain and at every $K$.

\begin{table}[t]
\centering
\caption{Reward discrimination on reference captions, the cells of
Figure~\ref{fig:ablation}(c): \textsc{cap}$\to$\textsc{ser} accuracy above chance,
$(\mathrm{acc}-1/K)/(1-1/K)$, for three verifiers and five candidate counts on the reference
captions of three domains.}
\label{tab:reader-grid}
\small
\setlength{\tabcolsep}{5pt}
\begin{tabular}{@{}llccccc@{}}
\toprule
Domain & Verifier & $K=4$ & $K=6$ & $K=8$ & $K=10$ & $K=12$ \\
\midrule
\multirow{3}{*}{TSFragment val.} & Qwen2.5-14B & 0.701 & 0.634 & 0.586 & 0.527 & 0.496 \\
 & Qwen2.5-7B & 0.641 & 0.553 & 0.491 & 0.443 & 0.401 \\
 & Qwen2.5-3B & 0.455 & 0.353 & 0.284 & 0.186 & 0.233 \\
\midrule
\multirow{3}{*}{CaTS-Bench} & Qwen2.5-14B & 0.421 & 0.384 & 0.357 & 0.336 & 0.324 \\
 & Qwen2.5-7B & 0.257 & 0.199 & 0.170 & 0.151 & 0.147 \\
 & Qwen2.5-3B & 0.112 & 0.100 & 0.095 & 0.080 & 0.105 \\
\midrule
\multirow{3}{*}{BEDTime} & Qwen2.5-14B & 0.194 & 0.165 & 0.155 & 0.142 & 0.146 \\
 & Qwen2.5-7B & 0.147 & 0.114 & 0.096 & 0.090 & 0.081 \\
 & Qwen2.5-3B & 0.038 & 0.041 & 0.031 & 0.032 & 0.036 \\
\bottomrule
\end{tabular}
\end{table}

\subsection{Per-setting ablation tables}
\label{app:results:ablation}

Tables~\ref{tab:abl_rm_caption}--\ref{tab:abl_sel_readout} give the per-benchmark and
per-corpus numbers behind Figure~\ref{fig:ablation}(a,b) and Table~\ref{tab:negrand}: for each
of the three ladders, the six caption-level metrics of Table~\ref{tab:criteria12} with the mean
caption length they were measured on, and the eight Criterion 3 cells of Table~\ref{tab:readout}.
The untuned initialization is repeated in every table as a reference line.
Table~\ref{tab:reader-grid} in Appendix~\ref{app:results:grid} lists the numbers behind Figure~\ref{fig:ablation}(c).

Two notes for reading these tables alongside the main ones. First,
the Criterion 3 numbers are single runs at predictor seed $2020$, where Table~\ref{tab:readout}
reports the mean over five, so the \method\ row reads $0.1687$ on forecasting ETTh2 here
against $0.1806$ there; across the five seeds of Table~\ref{tab:readout} the relative standard
deviation of a cell has median $3.6\%$ and $90$th percentile $6.1\%$, which is the scale below
which a difference in these tables should not be read. Second, on the reward-model ladder the
verifier that scores the reward changes with the setting, whereas the identification columns
are scored by the same two evaluation verifiers for every setting, so they compare the captions
on equal terms; entailment and Criterion 3 are not the objective of any setting.

\begin{table}[t]
\centering
\caption{Ablation on the reward model: caption metrics.
Effect of the size of the frozen verifier that turns a caption into the reward, with everything else held at the \method\ recipe.
Columns are those of Table~\ref{tab:criteria12}: \emph{Ent.} is the fraction of
reference captions the generated one entails, \textsc{c}$\to$\textsc{s} and
\textsc{s}$\to$\textsc{c} are the identification accuracies averaged over the two frozen
evaluation verifiers, GLM-4-9B and Phi-4-14B (chance $0.25$). \emph{Len.} is the mean caption
length in characters, printed because both entailment columns move with it. Bold marks the best
of the three trained settings; the untuned initialization is shown for reference. Figure~\ref{fig:ablation} plots the CaTS-Bench \textsc{s}$\to$\textsc{c} column.}
\label{tab:abl_rm_caption}
\small
\setlength{\tabcolsep}{4pt}
\begin{tabular}{@{}lcccccccc@{}}
\toprule
& \multicolumn{4}{c}{BEDTime} & \multicolumn{4}{c}{CaTS-Bench} \\
\cmidrule(lr){2-5} \cmidrule(lr){6-9}
Reward model & Ent. & \textsc{c}$\to$\textsc{s} & \textsc{s}$\to$\textsc{c} & Len. &
Ent. & \textsc{c}$\to$\textsc{s} & \textsc{s}$\to$\textsc{c} & Len. \\
\midrule
Untuned Qwen2.5-VL-3B & 0.296 & 0.550 & 0.495 & 1084 & 0.352 & 0.751 & 0.721 & 1287 \\
\midrule
Qwen2.5-14B (\textbf{ours}) & \textbf{0.400} & 0.709 & 0.638 & 621 & \textbf{0.611} & \textbf{0.886} & \textbf{0.868} & 713 \\
Qwen2.5-7B & 0.287 & 0.710 & 0.644 & 1331 & 0.298 & 0.850 & 0.799 & 1725 \\
Qwen2.5-3B & 0.270 & \textbf{0.715} & \textbf{0.710} & 898 & 0.381 & 0.830 & 0.831 & 959 \\
\bottomrule
\end{tabular}
\end{table}

\begin{table}[t]
\centering
\caption{Ablation on the reward model: Criterion 3.
Test MSE (lower is better) of the caption-only predictor of Table~\ref{tab:readout}, which
never sees the series. Every setting keeps $K=4$ candidates and the rest of the \method\ recipe; only the verifier that scores the reward changes.
\emph{Rel.} is the geometric mean of the four per-corpus ratios to the \method\ row, so
$>1$ means the caption carries less of what the predictor needs. Bold marks the best of the
three trained settings. Figure~\ref{fig:ablation} plots the \emph{Rel.} columns alone.}
\label{tab:abl_rm_readout}
\footnotesize
\setlength{\tabcolsep}{2.6pt}
\begin{tabular}{@{}lccccccccccc@{}}
\toprule
& \multicolumn{5}{c}{\textbf{Forecasting}} & \multicolumn{5}{c}{\textbf{Reconstruction}} \\
\cmidrule(lr){2-6} \cmidrule(lr){7-11}
Reward model & ETTh2 & ETTm2 & Saug. & Elec. & Rel. & ETTh2 & ETTm2 & Saug. & Elec. & Rel. \\
\midrule
Untuned Qwen2.5-VL-3B & 0.3114 & 0.1037 & 0.9234 & 0.5476 & 1.850 & 0.2469 & 0.0654 & 0.6698 & 0.4124 & 2.283 \\
\midrule
Qwen2.5-14B (\textbf{ours}) & \textbf{0.1687} & \textbf{0.0523} & \textbf{0.8225} & \textbf{0.1922} & 1.000 & \textbf{0.0974} & \textbf{0.0286} & \textbf{0.4472} & \textbf{0.1317} & 1.000 \\
Qwen2.5-7B & 0.2535 & 0.0758 & 0.8709 & 0.2926 & 1.369 & 0.1441 & 0.0435 & 0.5318 & 0.1847 & 1.391 \\
Qwen2.5-3B & 0.2312 & 0.0783 & 0.8459 & 0.3205 & 1.370 & 0.1693 & 0.0534 & 0.6552 & 0.2385 & 1.713 \\
\bottomrule
\end{tabular}
\end{table}

\begin{table}[t]
\centering
\caption{Ablation on the number of candidates: caption metrics.
Effect of the number of candidate series the verifier must choose among, with everything else held at the \method\ recipe.
Columns are those of Table~\ref{tab:criteria12}: \emph{Ent.} is the fraction of
reference captions the generated one entails, \textsc{c}$\to$\textsc{s} and
\textsc{s}$\to$\textsc{c} are the identification accuracies averaged over the two frozen
evaluation verifiers, GLM-4-9B and Phi-4-14B (chance $0.25$). \emph{Len.} is the mean caption
length in characters, printed because both entailment columns move with it. Bold marks the best
of the three trained settings; the untuned initialization is shown for reference. Figure~\ref{fig:ablation} plots the CaTS-Bench \textsc{s}$\to$\textsc{c} column.}
\label{tab:abl_neg_caption}
\small
\setlength{\tabcolsep}{4pt}
\begin{tabular}{@{}lcccccccc@{}}
\toprule
& \multicolumn{4}{c}{BEDTime} & \multicolumn{4}{c}{CaTS-Bench} \\
\cmidrule(lr){2-5} \cmidrule(lr){6-9}
Candidates & Ent. & \textsc{c}$\to$\textsc{s} & \textsc{s}$\to$\textsc{c} & Len. &
Ent. & \textsc{c}$\to$\textsc{s} & \textsc{s}$\to$\textsc{c} & Len. \\
\midrule
Untuned Qwen2.5-VL-3B & 0.296 & 0.550 & 0.495 & 1084 & 0.352 & 0.751 & 0.721 & 1287 \\
\midrule
$K=4$ (\textbf{ours}) & \textbf{0.400} & 0.709 & 0.638 & 621 & \textbf{0.611} & \textbf{0.886} & 0.868 & 713 \\
$K=6$ & 0.343 & 0.664 & 0.644 & 482 & 0.470 & 0.851 & 0.869 & 561 \\
$K=8$ & 0.340 & \textbf{0.790} & \textbf{0.796} & 555 & 0.391 & 0.852 & \textbf{0.876} & 620 \\
\bottomrule
\end{tabular}
\end{table}

\begin{table}[t]
\centering
\caption{Ablation on the number of candidates: Criterion 3.
Test MSE (lower is better) of the caption-only predictor of Table~\ref{tab:readout}, which
never sees the series. Every setting keeps the same frozen Qwen2.5-14B verifier and the rest of the \method\ recipe; only the number of candidates it must choose among changes. The $K-1$ distractors of a smaller $K$ are a prefix of those of a larger one, so raising $K$ adds candidates rather than exchanging them.
\emph{Rel.} is the geometric mean of the four per-corpus ratios to the \method\ row, so
$>1$ means the caption carries less of what the predictor needs. Bold marks the best of the
three trained settings. Figure~\ref{fig:ablation} plots the \emph{Rel.} columns alone.}
\label{tab:abl_neg_readout}
\footnotesize
\setlength{\tabcolsep}{2.6pt}
\begin{tabular}{@{}lccccccccccc@{}}
\toprule
& \multicolumn{5}{c}{\textbf{Forecasting}} & \multicolumn{5}{c}{\textbf{Reconstruction}} \\
\cmidrule(lr){2-6} \cmidrule(lr){7-11}
Candidates & ETTh2 & ETTm2 & Saug. & Elec. & Rel. & ETTh2 & ETTm2 & Saug. & Elec. & Rel. \\
\midrule
Untuned Qwen2.5-VL-3B & 0.3114 & 0.1037 & 0.9234 & 0.5476 & 1.850 & 0.2469 & 0.0654 & 0.6698 & 0.4124 & 2.283 \\
\midrule
$K=4$ (\textbf{ours}) & \textbf{0.1687} & 0.0523 & 0.8225 & \textbf{0.1922} & 1.000 & \textbf{0.0974} & 0.0286 & 0.4472 & 0.1317 & 1.000 \\
$K=6$ & 0.2046 & \textbf{0.0442} & \textbf{0.8067} & 0.1925 & 1.002 & 0.0975 & 0.0288 & \textbf{0.4179} & \textbf{0.1234} & 0.969 \\
$K=8$ & 0.2206 & 0.0529 & 0.8421 & 0.2120 & 1.106 & 0.1006 & \textbf{0.0271} & 0.4287 & 0.1382 & 0.996 \\
\bottomrule
\end{tabular}
\end{table}

\begin{table}[t]
\centering
\caption{Ablation on how the distractors are chosen: caption metrics.
Effect of the rule that picks the distractors an item is scored against, with everything else held at the \method\ recipe.
Columns are those of Table~\ref{tab:criteria12}: \emph{Ent.} is the fraction of
reference captions the generated one entails, \textsc{c}$\to$\textsc{s} and
\textsc{s}$\to$\textsc{c} are the identification accuracies averaged over the two frozen
evaluation verifiers, GLM-4-9B and Phi-4-14B (chance $0.25$). \emph{Len.} is the mean caption
length in characters, printed because both entailment columns move with it. Bold marks the best
of the two trained settings; the untuned initialization is shown for reference. Table~\ref{tab:negrand} aggregates these columns.}
\label{tab:abl_sel_caption}
\small
\setlength{\tabcolsep}{4pt}
\begin{tabular}{@{}lcccccccc@{}}
\toprule
& \multicolumn{4}{c}{BEDTime} & \multicolumn{4}{c}{CaTS-Bench} \\
\cmidrule(lr){2-5} \cmidrule(lr){6-9}
Distractors & Ent. & \textsc{c}$\to$\textsc{s} & \textsc{s}$\to$\textsc{c} & Len. &
Ent. & \textsc{c}$\to$\textsc{s} & \textsc{s}$\to$\textsc{c} & Len. \\
\midrule
Untuned Qwen2.5-VL-3B & 0.296 & 0.550 & 0.495 & 1084 & 0.352 & 0.751 & 0.721 & 1287 \\
\midrule
Nearest neighbor (\textbf{ours}) & \textbf{0.400} & \textbf{0.709} & \textbf{0.638} & 621 & \textbf{0.611} & \textbf{0.886} & \textbf{0.868} & 713 \\
Random & 0.324 & 0.598 & 0.636 & 723 & 0.384 & 0.810 & 0.853 & 753 \\
\bottomrule
\end{tabular}
\end{table}

\begin{table}[t]
\centering
\caption{Ablation on how the distractors are chosen: Criterion 3.
Test MSE (lower is better) of the caption-only predictor of Table~\ref{tab:readout}, which
never sees the series. Both settings keep the same frozen Qwen2.5-14B verifier, $K=4$ candidates and the rest of the \method\ recipe; only the rule that picks the $K-1$ distractors of a training item changes.
\emph{Rel.} is the geometric mean of the four per-corpus ratios to the \method\ row, so
$>1$ means the caption carries less of what the predictor needs. Bold marks the best of the
two trained settings. Table~\ref{tab:negrand} aggregates these columns.}
\label{tab:abl_sel_readout}
\footnotesize
\setlength{\tabcolsep}{2.6pt}
\begin{tabular}{@{}lccccccccccc@{}}
\toprule
& \multicolumn{5}{c}{\textbf{Forecasting}} & \multicolumn{5}{c}{\textbf{Reconstruction}} \\
\cmidrule(lr){2-6} \cmidrule(lr){7-11}
Distractors & ETTh2 & ETTm2 & Saug. & Elec. & Rel. & ETTh2 & ETTm2 & Saug. & Elec. & Rel. \\
\midrule
Untuned Qwen2.5-VL-3B & 0.3114 & 0.1037 & 0.9234 & 0.5476 & 1.850 & 0.2469 & 0.0654 & 0.6698 & 0.4124 & 2.283 \\
\midrule
Nearest neighbor (\textbf{ours}) & \textbf{0.1687} & \textbf{0.0523} & 0.8225 & \textbf{0.1922} & 1.000 & \textbf{0.0974} & 0.0286 & \textbf{0.4472} & \textbf{0.1317} & 1.000 \\
Random & 0.2314 & 0.0643 & \textbf{0.8138} & 0.2484 & 1.212 & 0.1397 & \textbf{0.0268} & 0.4513 & 0.1652 & 1.142 \\
\bottomrule
\end{tabular}
\end{table}

\clearpage

\section{Prompts}
\label{app:prompts}

Every prompt below is reproduced from the code that ran; typographic dashes and bullets are
rendered in ASCII and nothing else is changed.

\paragraph{Caption prompt.}\label{app:prompt:caption}
The single user turn every captioner in Tables~\ref{tab:criteria12} and~\ref{tab:readout} receives, at RL rollout, at SFT teacher generation, at validation and at evaluation. Number-reading models receive the same text with \emph{image}/\emph{chart} replaced by \emph{values}.
\begin{promptbox}
This image is a line chart of a single time series. Describe the series in detail: what values it takes, how it moves over time, and where along the time axis its notable features occur. Write it so that someone who cannot see the chart could tell this series apart from other series that look broadly similar.
\end{promptbox}

\paragraph{Verifier prompt of the \method\ reward.}\label{app:prompt:verifier}
System line followed by the user turn, at $K=4$. Every candidate is printed at the precision chosen for the true series, so formatting cannot single it out. Criterion~2 (\textsc{cap}$\to$\textsc{ser}) asks the same question of the evaluation candidates; \textsc{ser}$\to$\textsc{cap} swaps the two sides.
\emph{System:}
\begin{promptbox}
You are a careful time-series analyst. You answer with a single letter.
\end{promptbox}
\emph{User:}
\begin{promptbox}
Below is a written description of one time series.
\smallskip
Description:
"""\{caption\}"""
\smallskip
Here are four candidate time series, each given as its raw values in time order.
\smallskip
A) \{opt\_a\}
\smallskip
B) \{opt\_b\}
\smallskip
C) \{opt\_c\}
\smallskip
D) \{opt\_d\}
\smallskip
Exactly one of these four is the series that the description was written about.
Which one is it? Answer with a single letter: A, B, C, or D.
\end{promptbox}

\paragraph{Judge prompt of the LLM-as-judge reward.}\label{app:prompt:judge}
The judge receives the true series as raw numbers and the caption, never the chart or any other series, and returns an integer $1$--$10$.
\emph{System:}
\begin{promptbox}
You are a careful time-series analyst. You grade written descriptions of time series against the actual values. You answer with a score and nothing else.
\end{promptbox}
\emph{User:}
\begin{promptbox}
Here are the values of one time series, in order:
\smallskip
\{series\}
\smallskip
Here is a written description of that series:
\smallskip
\{caption\}
\smallskip
Grade the description on how well it describes THIS series, from 1 to 10.
\smallskip
What to weigh, in order:
\smallskip
1. Accuracy. Every claim it makes -- direction, shape, where features sit, magnitudes --
\ \ \ must be true of the values above. A confident claim that is wrong is the worst thing a
\ \ \ description can do and must pull the score below any description that says less.
2. Specificity. A description that would fit many different series is worth little, even
\ \ \ when nothing in it is false. Ask whether someone holding this description could pick
\ \ \ this series out from others of broadly similar shape.
3. Completeness. Whether the description covers the movements that matter, not whether it
\ \ \ covers everything.
\smallskip
Anchors:
\smallskip
\ \ 1-2   Mostly wrong, or so generic it would fit almost any series.
\ \ 3-4   Broadly right about the overall direction, wrong or silent on everything else.
\ \ 5-6   Correct about the main movement, vague about where features occur or how large
\ \ \ \ \ \ \ \ they are.
\ \ 7-8   Accurate and specific about the main features and roughly where they occur.
\ \ 9-10  Accurate throughout AND specific enough to identify this series among similar
\ \ \ \ \ \ \ \ ones.
\smallskip
Score 1 if the description is empty, or if it could have been written without seeing
these values at all. You are grading the description, not the series: when there is
nothing there to grade, the score is 1, however interesting the values are.
\smallskip
Do not reward length. A short description that is accurate and specific scores higher
than a long one that hedges, repeats itself, or pads with claims you cannot verify.
\smallskip
Reply with the integer score alone, no words, no punctuation.
\end{promptbox}

\paragraph{Question-writing prompt of the answerability reward.}\label{app:prompt:qagen}
System prompt and user turn given to the question-writing VLM, which sees the chart only; $N$ is filled in with the length of the series.
\emph{System:}
\begin{promptbox}
You are given a line chart of a single univariate time series. Generate TEN multiple-choice questions (with their answers) about THIS specific series.
\smallskip
Hard requirements:
- The ten questions must each probe a DIFFERENT aspect of the curve. Do not ask ten variants of the same thing, and do not always start with "where is the peak" or "what is the overall trend".
- Mix both kinds: some questions about the OVERALL behavior (trend, shape, how it starts vs. ends, which half/third is higher, steeper, or noisier) and some about a SPECIFIC salient event you can clearly see (a prominent peak, dip, surge, or plateau) -- when you ask about such an event, locate it by its coarse region and only if it is clearly visible.
- Each question must be SPECIFIC to this curve: refer to the concrete shape you actually see (a particular rise, dip, plateau, spike, oscillation, or where along the series it happens). Avoid generic, template-like questions that could apply to any series.
- Each question must be answerable purely from the curve, have exactly four options, and exactly one correct answer. Make the distractors plausible.
- Do NOT ask for exact numeric values, forecasting, or anything that is not visible in the chart.
- Even when the raw values are provided to you for reference, use them only to read the true shape -- do NOT ask for exact y-values or step-precise readings, and only ask about a feature (peak/dip/surge/plateau) if it is genuinely there in the values.
- Write every positional option as a PLAIN NUMERIC time index -- a SINGLE index like "t=7" when the answer is intrinsically one point, or an INTERVAL like "t=5-10" when the answer is a region/segment -- with NO verbal label (do not write "early part", "middle", etc.; those words are vague and ambiguous). The series has integer time indices 0 to N-1 (N is given to you).
- MATCH the option granularity to what the question actually asks -- do NOT force every option into a wide interval when that is the wrong shape:
\ \ \ - REGION/SEGMENT questions (overall trend, which part is steeper / higher-on-average / noisier, where a rise / dip / plateau spans): use COARSE intervals. Place boundaries where THIS curve actually changes so the four are generally UNEQUAL in width (do NOT just split into four equal quarters), each at least about one fifth of the series wide.
\ \ \ - SINGLE-POINT questions (where the exact maximum / peak or minimum / trough sits): use SINGLE time indices (or tight windows) as the options. Inflating a one-point answer into a wide interval makes several intervals tie and the question becomes unanswerable -- so be truthful and keep these options point-sized.
- CRITICAL -- the four options of a positional question must be MUTUALLY EXCLUSIVE (disjoint): no two may overlap or cover the same span/point (adjacent intervals may meet at a single boundary index only). If you cannot fill four disjoint options, make one option a NON-positional choice instead (e.g. "Roughly constant throughout / No single clear region").
\smallskip
Output format -- follow it EXACTLY. The example below shows ONLY the format; do NOT copy its wording or topic. Positional options are numeric and disjoint (intervals for region questions, single indices for single-point questions):
\#\#\#\# 1. **<a specific question about this curve>**
\ \ \ - A) **<Choice A>**
\ \ \ - B) **<Choice B>**
\ \ \ - C) **<Choice C>**
\ \ \ - D) **<Choice D>**
\smallskip
**Answer:** <letter>) <correct option text>
------
\smallskip
Produce questions 1 to 10 in this exact format, each block separated by a line containing only "------". Output nothing else.
\end{promptbox}
\emph{User:}
\begin{promptbox}
Here is the line chart of the time series. The series has N=\{n\} points (time indices 0..\{n-1\}). Write each positional option as a plain numeric time index or time interval only (e.g. 't=7' or 't=5-13'), with NO verbal region words. For region questions use coarse intervals (>= about N/5 wide) and place the boundaries where this curve actually changes so the four intervals are UNEQUAL in width; for single-point questions, such as where the peak or the trough sits, use single indices or tight windows. Keep the four options non-overlapping. Generate the ten questions now.
\end{promptbox}

\paragraph{Answerer prompt of the answerability reward.}\label{app:prompt:answerer}
The frozen text-only answerer is queried once per question in this chat template; the first letter in its reply is compared with the key.
\begin{promptbox}
<|im\_start|>user
You will be given a caption describing a univariate time series (a line chart of a value versus a time index).
Your task is to answer the multiple-choice question strictly based on the caption, even if the answer may seem obvious from prior knowledge or the question wording.
\smallskip
Ignore any outside knowledge. Do not assume anything the caption does not explicitly or implicitly state.
\smallskip
Example:
Caption: <Caption Start> The series rises steadily from about 10 at the start to a peak near 40 around the middle, then declines to roughly 25 by the end. <Caption End>
Question: At which point in time is the series at its highest value?
- A) the very beginning
- B) one quarter of the way through
- C) around the middle
- D) at the very end
\smallskip
The answer is C.
\smallskip
Now, answer the question based on the following caption:
\smallskip
Caption: <Caption Start> \{\} <Caption End>
Question: \{\}  <|im\_end|>
<|im\_start|>assistant
The answer is
\end{promptbox}

\paragraph{CaTS-Bench rewrite, pass 1 (domain removed).}\label{app:prompt:cats1}
Instruction given to each of the $30$ rewriting agents; the slice markers \texttt{\{\{START\}\}}, \texttt{\{\{END\}\}}, \texttt{\{\{K\}\}} and \texttt{\{\{NNN\}\}} differ per agent and \texttt{<work>} stands for the scratch directory of the run.
\begin{promptbox}
Each line of your slice is \{"id": "...", "gold": "<caption>"\}. The caption is your only input: there is no chart, no values and no metadata, and you must not go looking for any.
\smallskip
Rewrite each caption so it describes only what an unlabelled line chart could show. Take out the domain: the phenomenon or attribute name, any place, port, store or product name, units and currency (keep the bare number), comparisons to statistics outside this series (historical mean, all-time max, "the reference period"), and any explanation that needs to know what the data is.
\smallskip
Dates become relative positions, never counted ones. "From 2011 to 2018" becomes "across the series"; "peaking in August 2023" becomes "peaking about two-thirds of the way through"; "a dip in late May" becomes "a dip near the end". Do not say how many points the series has and do not number them -- no "the eighth point", no "40 points", no "index 6", and drop counts of time units such as "a five-week snapshot". The text does not tell you how many points there are, and a guess would put a false claim into the caption. The vocabulary is "at the start", "early on", "midway", "in the second half", "near the end", "about two-thirds of the way through".
\smallskip
Keep the same observations in the same order, and keep every value and statistic as written. Rephrase as freely as the removals require, so that what is left reads as fluent, natural prose rather than a sentence with holes in it. What you must not do is add an observation the caption did not make, or change a number, even if the caption looks wrong. If a sentence has nothing left after the removals, drop it.
\smallskip
Your slice: sed -n '\{\{START\}\},\{\{END\}\}p' <work>/items\_textonly.jsonl  (\{\{K\}\} lines)
Write <work>/shards/batch\_\{\{NNN\}\}.jsonl -- exactly \{\{K\}\} lines, same order, one JSON object per line built with python3 + json.dumps (never by hand), "id" copied unchanged, plus "rewritten". Rewrite them yourself, one by one; a script is only for reading input and writing output. Reply with the shard path and line count, nothing else.
\end{promptbox}

\paragraph{CaTS-Bench rewrite, pass 2 (numbers removed).}\label{app:prompt:cats2}
Applied to the output of pass~1, so the two reference sets differ only in the numerals.
\begin{promptbox}
Each line of your slice is \{"id": "...", "gold": "<caption>"\}. The caption is your only input: there is no chart, no values and no metadata, and you must not go looking for any.
\smallskip
Rewrite each caption so that it states the same observations with **no numbers at all**. The target is the register of these examples, which are real captions from another benchmark:
\smallskip
\ \ \ \ "line decreases near the end"
\ \ \ \ "highest point is in middle"
\ \ \ \ "rises in the middle then flattens"
\smallskip
Remove every numeral and every quantity: values ("starting at 91.29"), ranges, means, percentages, counts, and any word that only carries a magnitude ("15\%", "approximately 40 units"). Positions in time stay, because they are qualitative: "at the start", "early on", "midway", "in the second half", "near the end", "about two-thirds of the way through".
\smallskip
Replace a quantity with the comparative fact it supports, never with a vaguer number. "Starting at 91.29 and reaching 105.27" becomes "starting low and ending higher". "A notable increase of approximately 15\%" becomes "a clear increase". "Rising above 100 along the way" becomes "rising above its starting level along the way" ONLY if the caption itself says what 100 relates to; if it does not, drop that clause. "The mean of the series is 97.3" has no numberless form that says anything about THIS series against others -- drop it.
\smallskip
Keep the same observations in the same order, and keep every non-numeric detail: direction, shape, where features sit, whether movement is smooth or noisy, whether it is steady or abrupt. Do not add an observation the caption did not make. If a sentence is nothing but a quantity, drop the sentence. If dropping leaves the caption empty, write the single most specific shape statement the original supports.
\smallskip
The result must read as fluent prose, not as a sentence with holes in it, and must contain no digits.
\smallskip
Your slice: sed -n '\{\{START\}\},\{\{END\}\}p' <work>/items.jsonl  (\{\{K\}\} lines)
Write <work>/shards/batch\_\{\{NNN\}\}.jsonl -- exactly \{\{K\}\} lines, same order, one JSON object per line built with python3 + json.dumps (never by hand), "id" copied unchanged, plus "rewritten". Rewrite them yourself, one by one; a script is only for reading input and writing output. Reply with the shard path and line count, nothing else.
\end{promptbox}

\end{document}